\documentclass[10pt,twocolumn,letterpaper]{article}
\usepackage{iccv}
\usepackage{times}
\usepackage{epsfig}
\usepackage{graphicx}
\usepackage{amsmath, bm}
\usepackage{amssymb}

\usepackage[numbers,sort,compress]{natbib}
\usepackage[pagebackref=true,breaklinks=true,letterpaper=true,colorlinks,bookmarks=false]{hyperref}
\usepackage{latexsym}
\usepackage{url}
\usepackage{soul}
\usepackage{multirow}
\usepackage{xcolor}
\usepackage{enumerate}
\usepackage{xparse}
\usepackage{float,booktabs}
\usepackage{etoolbox}
\usepackage{cleveref}

\usepackage{mathtools}

\DeclarePairedDelimiter\floor{\lfloor}{\rfloor}

\usepackage[ruled,vlined]{algorithm2e}
\newcommand*{\argmin}{\operatornamewithlimits{argmin}\limits}
\newcommand*{\argmax}{\operatornamewithlimits{argmax}\limits}
\newcommand{\ctab}[1]{\begin{tabular}{c} #1 \end{tabular}}

\usepackage[toc,page]{appendix}

\definecolor{blue}{HTML}{4EBEFE}
\definecolor{themered}{HTML}{FF8375}

\NewDocumentCommand{\Narendra}{ mO{} }{\textcolor{themered}{\textsuperscript{\textit{Narendra}}\textsf{\textbf{\small[#1]}}}}
\NewDocumentCommand{\Xiaodan}{ mO{} }{\textcolor{blue}{\textsuperscript{\textit{Xiaodan}}\textsf{\textbf{\small[#1]}}}}

\iccvfinalcopy 

\def\iccvPaperID{10918} 

\ificcvfinal\fi

\begin{document}

\title{Unsupervised 3D Pose Estimation for Hierarchical Dance Video Recognition\thanks{The support of the Office of Naval Research under grant N00014-20-1-2444 and USDA National Institute of Food and Agriculture under grant 2020-67021-32799/1024178 are gratefully acknowledged.}}

\author{Xiaodan Hu, Narendra Ahuja\\
Department of Electrical and Computer Engineering\\
University of Illinois at Urbana-Champaign\\
{\tt\small \{xiaodan8,n-ahuja\}@illinois.edu}
}

\maketitle

\begin{abstract}
Dance experts often view dance as a hierarchy of information, spanning low-level (raw images, image sequences), mid-levels (human poses and bodypart movements), and high-level (dance genre). We propose a Hierarchical Dance Video Recognition framework (HDVR). HDVR estimates 2D pose sequences, tracks dancers, and then simultaneously estimates corresponding 3D poses and 3D-to-2D imaging parameters, without requiring ground truth for 3D poses. Unlike most methods that work on a single person, our tracking works on multiple dancers, under occlusions. From the estimated 3D pose sequence, HDVR extracts body part movements, and therefrom dance genre. The resulting hierarchical dance representation is explainable to experts. To overcome noise and interframe correspondence ambiguities, we enforce spatial and temporal motion smoothness and photometric continuity over time. We use an LSTM network to extract 3D movement subsequences from which we  recognize dance genre. For experiments, we have identified 154 movement types, of 16 body parts, and assembled a new University of Illinois Dance (UID) Dataset, containing 1143 video clips of 9 genres covering 30 hours, annotated with movement and genre labels. Our experimental results demonstrate that our algorithms outperform the state-of-the-art 3D pose estimation methods, which also enhances our dance recognition performance.

\end{abstract}

\section{Introduction}
\label{sec:1introduction}


\begin{figure*}[!ht]
\centering
\begin{tabular}{c}
\includegraphics[width=17cm,keepaspectratio]{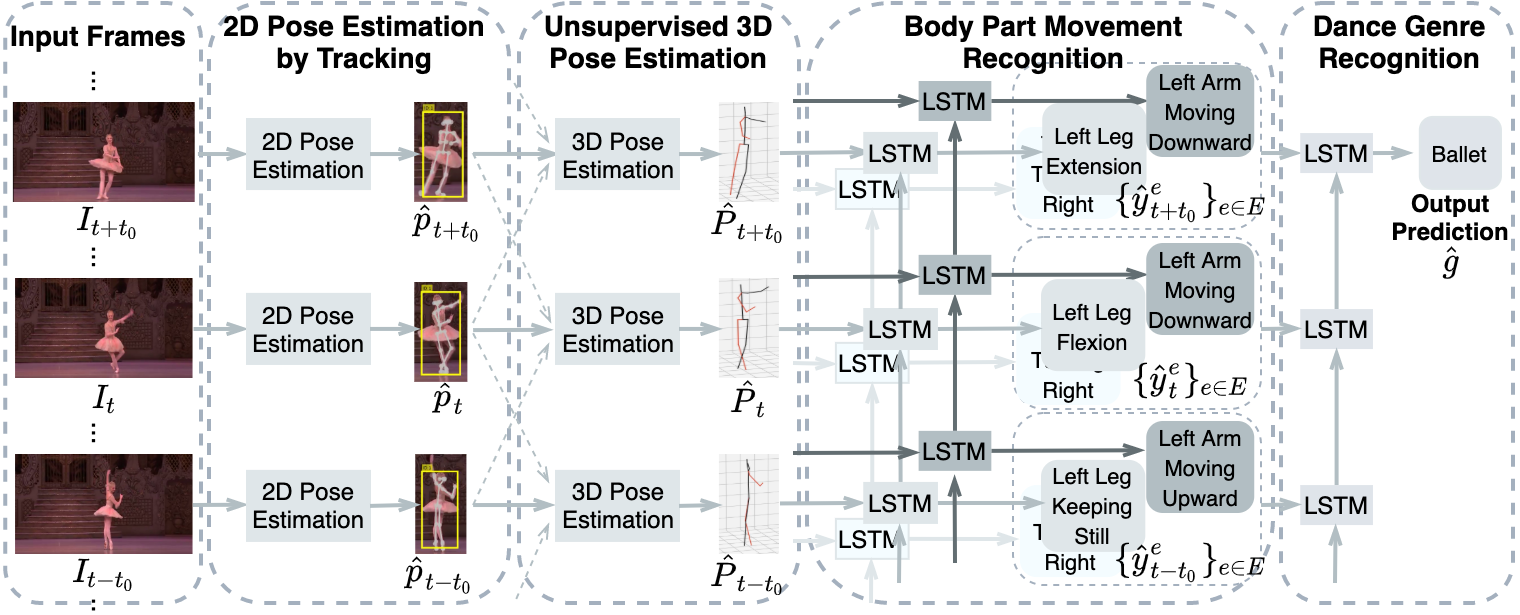}
\end{tabular}
\caption{
Overview of the model architecture. Given a sequence of video frames $\{I_t\}^{T-1}_{t=0}$, the model analyzes the content in a hierarchical manner, from the low levels (pose estimation \& tracking) to the cognitive levels (movement and dance genre recognition). The input sequence $\{I_t\}^{T-1}_{t=0}$ forms the first (bottom) level. At the second level, our algorithm simultaneously estimates the 2D pose $\hat{p}^i_t$ and 3D pose $\hat{P}^i_t$ of each dancer $i (i=0,...,N-1)$ at each frame, as well as the camera projection parameters. Our algorithm works under occlusions, e.g., among dancers.
At the third level, each dance movement $\hat{y}^e_t$ of each body part $e \in E$ (defined over a sequence of frames) is recognized and its location, given by, e.g., its starting frame $t$ and length are estimated, based on the poses estimated for the previous frames. At the fourth level, the dance genre $\hat{g}$ is recognized based on the movements $\{\hat{y}^e_t\}_{e\in E}$ of all body parts.  
}
\label{fig:overview}
\vspace{-0.2cm}
\end{figure*}


Dance represents a special genre of human activity. Our goal in this paper is development of algorithms to understand dance videos. We combine estimation of body movements with their feasibility as a part of dance. This enables interpretation of dance videos using not only constraints posed by the data but also those by the domain knowledge. 


A variety of proposed methods have also focused on dance videos \cite{Protopapadakis,Matsuyama,Dewan2017ICMV,Dewan2018laban,Castro2018,Zhao2019CVPR}. Most of these rely on kinect sensors to obtain depth information \cite{Protopapadakis,Matsuyama}. \cite{Dewan2017ICMV} classifies Indian dances by extracting patches centered at body's joint locations and using an LSTM network for classification. \cite{Dewan2018laban} proposes to
perform Laban Movement Analysis (in terms of dance domain constructs of Body, Effort, Shape and Space) to then describe human motion from a pose sequence.
\cite{Castro2018} compares the effects of using three different representations - raw images, optical flow and multi-person pose data - on their proposed dance dataset {proving that visual information is not sufficient to classify motion-heavy categories}. 
{
There are several approaches to action recognition that first estimate poses \mbox{\cite{Wang2013cvpr,Wang2019,Luvizon2018}}.
\mbox{\cite{Wang2013cvpr}} creates a coaching system for personalized athletic training based on pose correctness.
\mbox{\cite{Wang2019}} improves action recognition performance by improving pose estimation accuracy using additional spatial and temporal constraints.
However, \mbox{\cite{Wang2013cvpr,Wang2019}} both estimate only the 2D poses, leading to difficulties ambiguity when the movements are along the viewing direction .
\mbox{\cite{Luvizon2018}} estimates both 2D and 3D Poses as well as image features to predict actions from all three.
\mbox{\cite{Wang2013cvpr,Wang2019,Luvizon2018}} limit their representation for action recognition to pose sequences without including any higher level semantics that may define action. Moreover, these methods also require pose annotations in training videos.
}
\cite{Zhao2019CVPR} embeds RGB and optical-flow values into a single two-in-one stream network
for more efficient dance genre classification. 
In addition to the features such as pose and optical flow used in these works, in this paper we use dance domain representations to tune feature analysis to dance instead of being generic.

{When people dance, they follow a carefully choreographed sequence of 3D \textit{movements}, where each movement is hierarchically composed of simpler movements, ending in \textit{basic movements}. Each basic movement is composed of a sequence of poses representing a specific dance pattern. For brevity, in what follows, we will refer to basic movements by simply movements, 
We identify movements of 16 main \textit{body parts} $e \in E$ illustrated in Figure \mbox{\ref{fig:human_model}}. following Labanotation \mbox{\cite{labanotation}}, a well-known notation system used to record and archive human motion. Then in Table \mbox{\ref{tab:labels_movement}} we list the basic \textit{movements} $y^e \in Y^e$ for each \textit{body part} $e \in E$, again following \mbox{\cite{labanotation}} and defined in terms of homogeneity of motion direction, and level which are frequently used to describe the dance in dance domain.}
Our dance recognition model adopts this hierarchy used by dance experts, which starts with the 3D pose sequence of the dancer, combines subsequences of joint displacements into dance \textit{movements}, and finally infers dance genre from the sequences of the movements of joints. {To help the model segment the pose sequence into the basic movements, we manually annotate the starting and ending positions of such movements for each body part for a subset of videos in the UID dataset.}
Our framework takes a raw dance video sequence $\{I_t\}^{T-1}_{t=0}$ as input, estimates poses $\hat{p}_t$ for each frame $I_t$, recognizes the movement $\hat{y}^e_t$ (over multiple frames) of each body part $e$ based on its past pose sequence, and then predicts the dance genre $\hat{g}_t$ from the movement sequence. Experiments show that our hierarchical feature analysis is an effective way to recognize dance and our method outperforms state-of-the-art on F-score.

The main contributions of this paper are as follows:
\begin{itemize}
\item We propose the first dance video understanding framework that analyzes the videos hierarchically - from the bottom level of video frames, through the middle level of human poses, to the highest level of movements and associated dance genres.
\item Our algorithm tracks and outputs 2D pose of each dancer in each frame in the presence of occlusions among dancers.
\item We propose an unsupervised 3D pose estimation algorithm that starts with the estimated 2D pose sequence, and simultaneously and iteratively updates 2D poses, 3D poses and 3D-to-2D projection parameters 
using a single camera without using ground-truth for these poses or parameters.
Our 3D pose network achieves state-of-the-art performance by incorporating kinematic constraints of a 34-DOF human skeletal model and temporal smoothness of motion.
\item We have curated a large dance video data set, containing pose in ground truths for each video frame as well as for each movement,
which we will share with the community for further exploration.
\end{itemize}
\section{Computational Approach}

\begin{table*}[!ht]
\small
\centering
\begin{tabular}{l p{13.2cm} l}
	\toprule
	\textbf{Body Part} & \textbf{Examples of Movement Label} & \textbf{\# Labels} \\
	\midrule
    Head & Head Turning Up; Head Turning Down; Head Turning Left; Head Turning Right; Head Circling & 7 \\
    Neck & Neck Moving Left; Neck Moving Right; Neck Circling; Head Keeping Still; Unknown & 5 \\
	Left Shoulder & Left Shoulder Moving Upward; Left Shoulder Moving Downward; Left Shoulder Circling & 5 \\
	Left Lower Arm & Left Arm Moving Upward; Left Arm Moving Downward; Left Arm Moving Left & 11 \\
	Left Upper Arm & Left Arm Moving Upward; Left Arm Moving Downward; Left Arm Moving Left & 11 \\
	Torso & Torso Bending; Torso Unbending; Torso Turning Left; Torso Turning Right; Torso Swing; Somesault & 10 \\
	Hips & Hips Waving; Hips Figure 8; Hips Circling; Hip Moving Up; Hip Moving Down; Hips Keeping Still  & 10 \\
	Left Lower Leg & Left Leg Moving Upward; Left Leg Moving Downward; Left Leg Moving Left & 15 \\
	Left Upper Leg & Left Leg Moving Upward; Left Leg Moving Downward; Left Leg Moving Left & 15 \\
	Left Foot & Left Foot Extension; Left Foot Flexion; Left Foot Relaxed; Unknown & 4 \\
	\bottomrule
\end{tabular}
\caption{Selected examples of movement labels of each body part. To save space, only the movements of the left body parts are shown in the table. The movements of the right body parts are the same as the left ones. There are 16 body parts and 154 movement labels in total.
}
\label{tab:labels_movement}
\vspace{-0.2cm}
\end{table*}

Figure \ref{fig:overview} describes the components of our approach to dance video recognition and the hierarchy they form. Our approach can be summarized in the following steps:
Step 1: For each input frame $I_t$, the model estimates the 2D pose $p^i_t$ for 
dancer $i$ appearing in $I_t$. The model tracks approximate locations of the dancers $\{i\}^{N-1}_{i=0}$ throughout the video via their bounding boxes $\{B_t^i\}^{T-1}_{t=0}$. Step 2: At each frame, the model provides an estimate $\hat{p}^i_t$ of the 2D pose $p^i_t$ of the dancer associated with each tracked box $B_t^i$ (Section \ref{sec:2Dpose}). Step 3: {The model then estimates 3D poses $\hat{P}^i_t$ from the estimated 2D ones $\hat{p}^i_t$, by using an unsupervised 3D pose estimation method (Section \mbox{\ref{sec:3Dpose}}).} Step 4: The model uses the LSTM network to recognize the movement $\{\hat{y}^e_t\}^{T-1}_{t=0}$ of each body part $e \in E$ (e.g., head, torso, etc.) from the trajectories $\{\{\hat{P}^j_t\}_{j\in J_e}\}^{T-1}_{t=0}$ of all the joints $j \in J_e$ connected to the body part $e$, where $J_e \subset E$ (Section \ref{sec:Movement}). We represent any given state of a dance as a set of body part configurations and the entire dance as a sequence of such sets. Step 5: For recognition, we first concatenate the movements $\{\{\hat{y}^e_t\}_{e\in E}\}\}^{T-1}_{t=0}$ of all body parts, and input it to an LSTM network to recognize the dance genre $\hat{g}$ (Section \ref{sec:Genre}). The rest of this section introduces the components of this hierarchy.


\subsection{2D Pose Estimation by Tracking}
\label{sec:2Dpose}

\begin{algorithm}[!ht]
\SetAlgoLined
\textbf{Input}: a sequence of video frames $\{I_t\}^{T-1}_{t=0}$ \\
\textbf{Output}: a sequence of bounding boxes $\{(x_t^i, y_t^i, w_t^i, l_t^i)\}^{T-1}_{t=0}$ of the $i^{th}$ dancer \\
Initialization: select the bounding box $(x_0^i, y_0^i, w_0^i, l_0^i)$ of $N$ dancers to track by mouth \\
\While{new frame $I_t$ available}{
\For {$i^{th}$ dancer}{ 
Obtain $(x_t^i, y_t^i, w_t^i, l_t^i)$ by LDES approach \\
\If{not overlap with others}{
  Store histogram and velocity of $i^{th}$ dancer \\
  }
\If{overlap happens \& tracking fails}{
  Estimate when overlap ends
  }
\If{overlap ends}{
  Relocate the bounding box
  }
}
}
\caption{Object Tracking}
\label{alg:tracking}
\end{algorithm}

\begin{algorithm}[!ht]6

\SetAlgoLined
\textbf{Input}: a sequence of video frames $\{I_t\}^{T-1}_{t=0}$ and a sequence of bounding boxes $\{B_t^i\}^{T-1}_{t=0} = \{(x_t^i, y_t^i, w_t^i, l_t^i)\}^{T-1}_{t=0}$ of the $i^{th}$ dancer\\
\textbf{Output}: a sequence of poses $\{\hat{p}_t^i\}^{T-1}_{t=0}$ of the $i^{th}$ dancer \\
\While{new frame $I_t$ available}{
Estimate poses // Perform OpenPose\\
\For {$i^{th}$ dancer}{ 
Select pose $\hat{c}$ from $C$ poses overlapped with the bounding box $B_t^i$ based on histogram match
}
}
\caption{Tracking Based 2D Pose Estimation}
\label{alg:2Dpose_est}
\end{algorithm}

To estimate 2D (or 3D) pose, we estimate 2D (or 3D) coordinates of each body joint. Classical pose estimation methods such as pictorial structures framework and deformable part models largely rely on hand-designed features to determine body joint locations.
{Recently, deep learning-based approaches have achieved a major breakthrough in solving the problems in multi-person pose estimation (e.g., how to group keypoints for different people). They
can be divided into top-down \mbox{\cite{Wang2020, 2018PersonLab}} and bottom-up \mbox{\cite{PifPaf2019,HigherHRNet2020,OpenPose}}.}
The former employ detectors to first locate person instances and then their individual joints; the latter first estimate all joint locations within the image and then assign the joints to the associated person. 
Although these methods provide superior pose estimates, they have two major shortcomings critical to our task. Firstly,
most of the pose estimation methods cannot track a dancer through the video when there are multiple dancers present 
because they perform pose estimation from individual images, ignoring the temporal information. 
Besides, the methods perform training mostly on large datasets wherein the dance parts are very small,
with a single person, limited pose variety, and clean background.
and therefore cannot guarantee accuracy on real world dance videos.
The method we propose
can track selected dancers, detect estimation errors, and correct them automatically.

\begin{figure*}[!ht]
\centering
\begin{tabular}{c}
\includegraphics[width=17.2cm,keepaspectratio]{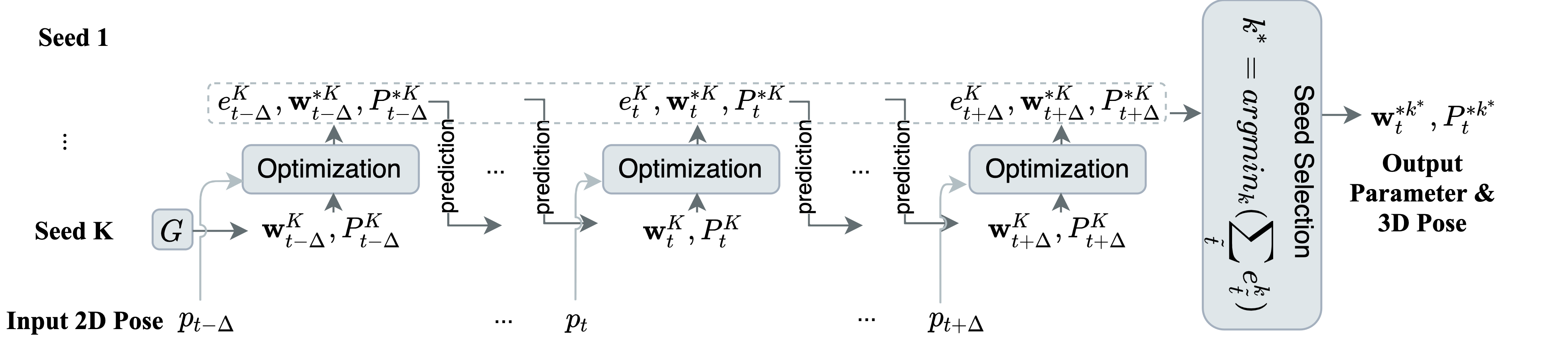} \\
\includegraphics[width=17.2cm,keepaspectratio]{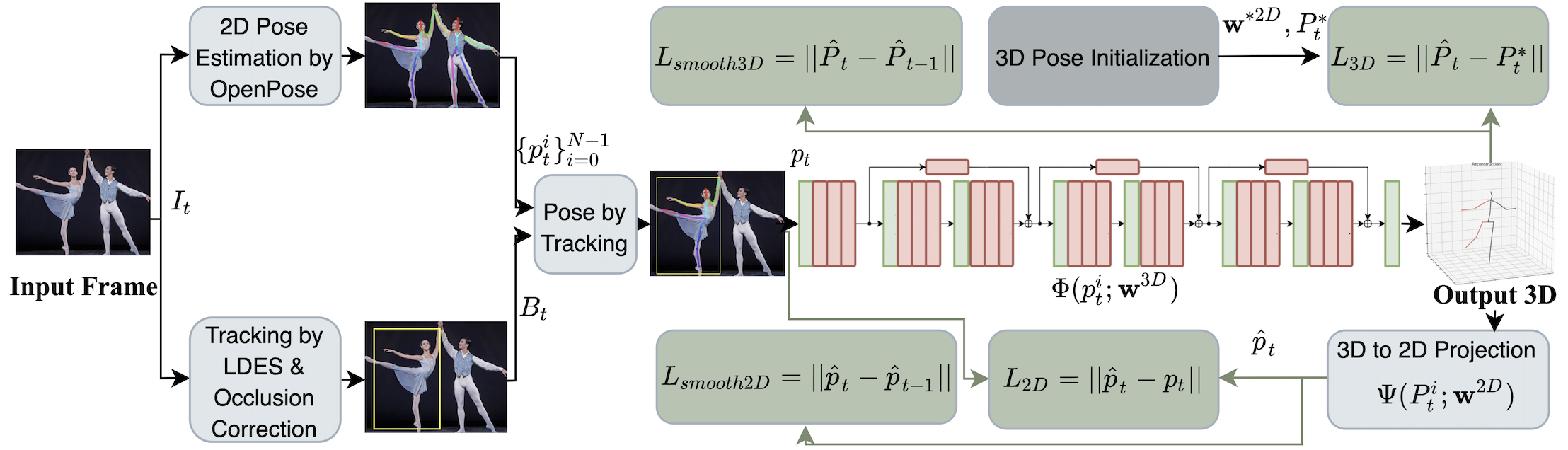}
\end{tabular}
\caption{Overview of Proposed 3D pose estimation method. Given a sequence of video frames $\{I_t\}^{T-1}_{t=0}$, the dancers are tracked by our tracking algorithm in Algorithm \mbox{\ref{alg:tracking}} and each of their 2D poses $\{p_t^i\}^{N-1}_{i=0}$ are estimated by our tracking based 2D pose estimation algorithm in Algorithm \mbox{\ref{alg:2Dpose_est}}. Then based on the 2D poses $\{p_t^i\}^{N-1}_{i=0}$, we initialize their 3D poses and camera perspective projection parameters, $P_{t}^{*}$ and $\omega^{*2D}$,
as shown in Fig. \mbox{\ref{fig:3Dpose} (top)}
and Algorithm \mbox{\ref{alg:3D_to_2D}}. Finally,
a neural network is trained to estimate the 3D poses $\{\hat{P}_t\}^{T-1}_{t=0}$, which incorporates kinematic constraints and spatiotemporal smoothness of motion, as described in Algorithm \mbox{\ref{alg:3D}}.}
\label{fig:3Dpose}
\vspace{-0.5cm}
\end{figure*}

\noindent\textbf{Object Tracking:} As explained in Algorithm \ref{alg:tracking}, our tracking algorithm is built upon the LDES tracker \cite{Li2019LDES}. Since occlusion between dancers is a serious problem, our algorithm centrally addresses it. Following are the three stages of our algorithm: (1) Use the LDES tracker to track each $i^{th}$ dancer when the dancer has no overlap with other dancers, while maintaining a color histogram $h^i_t$ and a bounding box $B^i_t=(x_t^i, y_t^i, w_t^i, l_t^i)$ for the dancer.
(2) Detect occurrence of overlap by detecting failure of the tracker as indicated by 
a significant difference between the directions of motion before and after overlap.
(3) 
Predict the time and the location of the dancer when the overlap may be expected to end, from the location and velocity observed just before the beginning of the overlap. Since multiple dancers may be detected in the vicinity of the predicted location in the predicted frame, select the one that provides the best histogram match, and update $h^i_t$ and $B^i_t$ accordingly.

\noindent\textbf{Tracking Based 2D Pose Estimation:} 
{
As explained in Algorithm \mbox{\ref{alg:2Dpose_est}}, we obtain the initial 2D poses by using the OpenPose method \mbox{\cite{OpenPose}}.
}
After we obtain the bounding box $B^i_t$ for each dancer $i$ at the end of the overlap, the box $B^i_t$ may overlap with multiple boxes simultaneously, indicating multiple 2D pose estimation results.
We select that pose $\hat{p}^i_t$ whose histogram is most similar to the one $\hat{p}^i_{t-1}$ seen in the previous frame. (Algorithm \mbox{\ref{alg:2Dpose_est}}).

\subsection{3D Pose Estimation}
\label{sec:3Dpose}

\begin{algorithm}[!ht]
\SetAlgoLined
\textbf{Input}: a sequence of 2D poses $\{p_t\}^{N-1}_{t=0}$ of a dancer \\
\textbf{Output}: a sequence of 3D poses $\{\tilde{P}_t\}^{N-1}_{t=0}$ of the dancer \\
Set the temporal window size to be $2\Delta$ \\
Denote total number of segments as $s=\floor*{\frac{N}{2\Delta}}$ \\
\For{$t = \Delta$ to $N-\Delta$}{
\For{k = 0 to $K-1$}{
Try new seed for DH parameters $\Lambda^k$ and perspective projection parameters $\omega^k$ \\
\For{$i = t-\Delta$ to $t+\Delta$}{
Generate 3D pose $\hat{P}^k_i=G(\Lambda^k)$ \\
Estimate 2D pose $\hat{p}_{i}^k=\Psi(\hat{P}_{i}^k; \omega^k)$ \\
Compute error $e^k_i = ||\hat{p}_{i}^k - p_{i}||^2_2$ \\
Optimize $\Lambda^{*k}, \omega^{*k}$ \\
}
}
}
Select the 3D pose corresponding to seed $k^{*} = \argmin_{\tilde{k}}\sum_{i=t-\Delta}^{t+\Delta} e^{\tilde{k}}_i$
as the initialized pose\\
\caption{3D Pose Initialization}
\label{alg:3D_to_2D}
\end{algorithm}

\begin{algorithm}[!ht]
\SetAlgoLined
\textbf{Input}: a sequence of video frames $\{I_t\}^{T-1}_{t=0}$, 2D poses $\{p_t\}^{T-1}_{t=0}$ and initial 3D poses $\{\tilde{P}_t\}^{T-1}_{t=0}$ of a dancer \\
\textbf{Output}: a sequence of estimated 3D poses $\{\hat{P}_t\}^{T-1}_{t=0}$ of the dancer \\
\While{new frame $I_{t}$ available}{
Estimate 3D pose $ \hat{P}_{t}$ \\
Project to 2D pose $\hat{p}_{t}$ \\
Compute loss $L = \alpha(|| \hat{p}_{t} - \hat{p}_{t-1} ||^2_2 + \beta || \hat{P}_{t} - \hat{P}_{t-1} ||^2_2) + || \hat{p}_{t} - p_{t} ||^2_2 + || \hat{P}_{t} - \tilde{P}_{t} ||^2_2$ \\
Update $\omega^{2D}$ and $\omega^{3D}$
}
\caption{3D Pose Estimation}
\label{alg:3D}
\end{algorithm}

Towards our objective of using dance representations close to those used by experts, we need to use 3D, instead of 2D, pose sequences. Similarly for recognition using the language of dance experts, we need to extract descriptors of 3D movements from the 2D pose sequences, which constitute our method's next stage. Computationally too, 3D poses contain more information than 2D poses, and thus lead to more accurate dance recognition. However, predicting 3D poses from 2D poses is an ill-posed problem like other 2D-to-3D problems. The state-of-the-art methods \cite{Pavllo2019CVPR,Wang20203D,Kocabas2020} use a two-step pipeline for solving it: first detect 2D poses from video frames, and then predict 3D poses 
by learning the correspondences of 2D and 3D key points.
{
\mbox{\cite{Martinez2017ICCV}} provides a simple yet effective baseline proving that the 2D to 3D task can be solved with a remarkably low error rate.
\mbox{\cite{Wandt2019RepNet}} learns a mapping from a distribution of 2D poses to a distribution of 3D poses using
an adversarial training approach. However, \mbox{\cite{Martinez2017ICCV,Wandt2019RepNet}} estimate 3D poses from 2D poses estimated from individual 2D frames, which ignores the temporal continuity information.
}
\cite{chen2020anatomyaware,Xu2020cvpr} use temporal correspondences of 2D keypoints to both learn the joint angles as well as predict the joint locations.
They compute loss in terms of the distance between these key points and those back-projected using the estimated 3D pose. They enforce such geometric consistency to progressively refine the estimates of 3D poses. 
However, these methods are based on the assumption that the input 2D poses are accurate. 
\cite{Xu2020cvpr} proposes a 2D pose correction module which uses a temporal CNN to refine the 2D initial inputs. However, this assumes that ground-truth 2D poses are available to train the correction module. These assumptions are often restrictive in practice, and do not hold for our dance videos which are collected from the internet.
{
\mbox{\cite{Andriluka2010Mono}} relates detected 2D poses across frames based on tracking-by-detection and then recovers 3D pose in a Bayesian framework. However, their MAP estimation is not robust if the video is long or background changes dramatically.
\mbox{\cite{XNect2020SIGGRAPH}} proposes a method to cope with occlusion. They first infer 3D locations of the visible body joints and then reconstruct the occluded joint locations using learned pose priors and a kinematic skeletal model. 
\mbox{\cite{Zanfir2018CVPR}} fit a parametric human model (SMPL) to observed image key points and segments along with some additional constraints.
However, \mbox{\cite{XNect2020SIGGRAPH,Zanfir2018CVPR}} require 3D pose labels and/or shape to supervise the training, which are not available for our ``in the wild'' video dataset.
\mbox{\cite{Zhou2017ICCV,kocabas2019epipolar}} estimate 3D pose from in-the-wild images without 3D pose annotations, but they require either additional 2D pose datasets or a multi-view setting.
}
To avoid these requirements and the need for groudtruth 2D pose, and to improve computational robustness, we propose an algorithm that integrates 3D pose estimation with 2D pose correction, which can be trained to converge on both estimates simultaneously while also estimating the camera projection parameters consistently.

\begin{figure}[!ht]
\centering
\begin{tabular}{c}
\includegraphics[width=8cm,keepaspectratio]{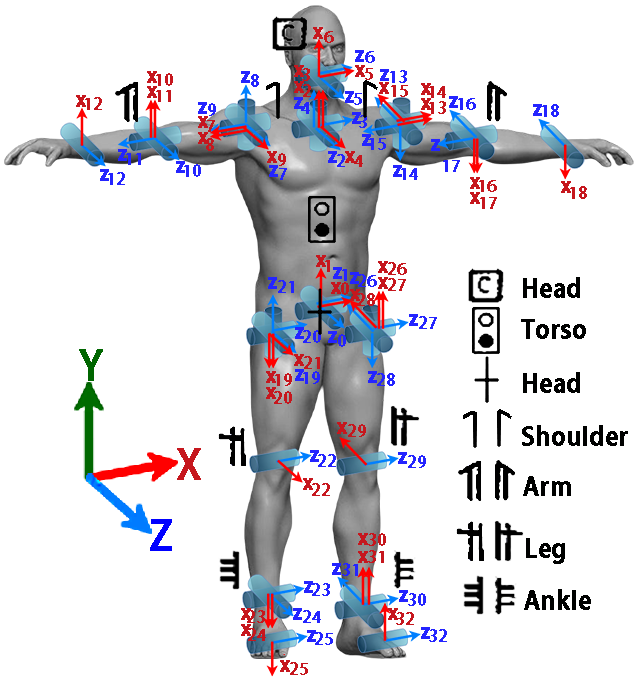}
\end{tabular}
\caption{Our 34-DOF digital dancer model. The values of the DH parameters $\Lambda = \{\bm{\Theta},\bm{d},\bm{a},\bm{\alpha}\}$ of this model are listed in Table \mbox{\ref{tab:DH_parameters}} 
in Appendix. The bounds of the joint rotation offset angles $\bm{\theta}$ and bone length $\bm{b}$ are defined in Table \mbox{\ref{tab:bounds}} 
in Appendix.}
\label{fig:human_model}
\end{figure}



{We use the Denavit-Hartenberg (DH) parameters $\Lambda^k = \{\bm{\Theta}^k,\bm{d}^k,\bm{a}^k,\bm{\alpha}^k\}$ to represent the 3D pose. A 3D pose $\tilde{P}_{t}$ is generated by passing $\Lambda^k$ to the 34-DOF kinematic model $G$ as follows:}
\begin{equation}
\hat{P}_{i}^k=(J_0, J_1, ..., J_{24})
\label{eq:P}
\end{equation}
\begin{equation}
J_j = G(\Theta, d, a, \alpha)=\mathcal{T}_{\Theta} \mathcal{T}_d \mathcal{T}_a \mathcal{T}_{\alpha} J_{j-1}
\label{eq:J}
\end{equation}
where
\begin{align*}
\scriptsize
\mathcal{T}_{\Theta} \mathcal{T}_d \mathcal{T}_a \mathcal{T}_{\alpha} =
\left[\begin{array}{cccc}
\cos \Theta & -\sin \Theta \cos \alpha & \sin \Theta \sin \alpha & r \cos \Theta \\
\sin \Theta & \cos \Theta \cos \alpha & -\cos \Theta \sin \alpha & r \sin \Theta \\
0 & \sin \alpha & \cos \alpha & d \\
0 & 0 & 0 & 1
\end{array}\right]
\end{align*}
where $\mathcal{T}_{\Theta}, \mathcal{T}_d, \mathcal{T}_a$ and $\mathcal{T}_{\alpha} J_{j-1}$ are transition matrices, and $J_j$ is the 3D location of the joint $j$.

We initialize the desired estimates of 3D pose $\tilde{P}_{t}$ and the 3D-to-2D projection parameter $\omega^{2D}_t$ with multiple randomly selected seed pairs $\{\Lambda^{*k}, \omega^{k}\}$ (to sample the search space), as explained in Figure \mbox{\ref{fig:3Dpose} (top)} and Algorithm \mbox{\ref{alg:3D_to_2D}}. $\omega^k=\{f^k,c^k\}$ are the perspective projection parameters. At frame $t$, we sample $K$ seeds of the DH parameters to generate 3D poses $\{\hat{P}_{i}^k\}_{i=t-\Delta}^{t+\Delta}$ in a sliding window of size $2\Delta$ centered at $t$ {and 3D-to-2D projection parameter $\omega^{2D}$}. 
By comparing the reconstructed 2D pose $\hat{p}=\Psi(\hat{P}_{t}; \omega^{2D})$ projected from the generated 3D pose $\hat{P}_{t}$
with the input 2D pose $p_{i}$ estimated in \mbox{\ref{sec:2Dpose}}, we optimize the DH parameters $\Lambda^k$ generating the 3D pose $\hat{P}_{i}^k$ while enforcing: (a) constraints that govern the joint rotation offset angles $\bm{\theta}^k$, (b) consistency with the known bone lengths $\bm{b}^k$
and (c) temporal smoothness of both the 2D and 3D poses. This is achieved by training with a loss function consisting of two parts: (1) temporal smoothness of both the 2D pose and 3D pose: $\alpha(|| \hat{p}_{t} - \hat{p}_{t-1} ||^2_2 + \beta || \hat{P}_{t} - \hat{P}_{t-1} ||^2_2)$. (2)  preservation of 3D-to-2D projection (imaging) property:
$|| \Psi(\hat{P}_{t}; \omega^{2D}) - p_{t} ||^2_2$. The coefficients $\alpha$ and $\beta$ are chosen to be inversely proportional to the error: the larger the error, smaller the weight of the window. 
We also enforce constancy of the 3D to 2D projection parameters by smoothing it over a time window.
At each time step $t$, we update the 3D pose $\hat{P}_{t}$ and the projection parameter $\omega^{3D}$. From among the solutions obtained using the different seeds, the pair $\{\hat{P}^{*}_{t}; \omega^{*2D}\}$ corresponding to the seed offering the least error is selected.


As shown in Figure \mbox{\ref{fig:3Dpose}} (bottom), after obtaining the initial 3D pose $P^{*}_{t}$ and the 3D-to-2D projection parameters $\omega_t^{*2D}$ from the \textit{3D Pose Initialization} block, we train temporal 
convolutional networks to learn the mapping from the input 2D poses $\{\hat{p}_t\}$ to the 3D ones $\{\hat{P}_t\}$. We use \mbox{\cite{Pavllo2019CVPR}} as our baseline networks.
During the training, in addition to the consistency between 2D and 3D poses at all times, we again enforce temporal smoothness of motion with the loss function defined as follows:
\begin{equation}
\mathcal{L} = || \hat{p}_{t} - p_{t} ||^2_2 + || \hat{p}_{t} - p_{t-1} ||^2_2 + || \hat{P}_{t} - P^{*}_{t} ||^2_2 + || \hat{P}_{t} - \hat{P}_{t-1} ||^2_2)
\label{eq:L}
\end{equation}
where $\hat{p}_{t}=\Psi(\hat{P}_{t}; \omega^{*2D})$.See details in Algorithm \ref{alg:3D}.

{
To further improve the accuracy when limited labeled 3D ground-truth pose data are available, we introduce a semi-supervised training version of the proposed pose estimation method. 
A supervised loss is trained by using the available labeled ground truth 3D poses $P_t$ as target, and the loss in Equation (\mbox{\ref{eq:L}}) is implemented using the remaining unlabeled data. Here, the predicted 3D poses $\hat{P}_t$ are projected back to 2D joint coordinates for consistency with the 2D input $p_t$. Similar to the training strategy in \mbox{\cite{Pavllo2019CVPR}}, we jointly optimize the supervised component with our unsupervised component during training, with the labeled data occupying the first half of a batch, and the unlabeled data occupying the second half.
}

\subsection{Body Part Movement Recognition}
\label{sec:Movement}

For each body part $e$, we train an LSTM-based model to recognize its (basic) movement. During training, the input is a sequence of 3D poses $\{\{\hat{p}^j_t\}_{j\in J_e}\}^{T-1}_{t=0}$ of all the joints $j \in J_e$ connected to the body part $e$ and the output is a sequence of predicted movement labels $\{\hat{y}^e_t\}^{T-1}_{t=0}$ connected to $e$. Since this is a multi-label classification problem, which means the poses $\{\hat{p}^j_t\}_{j\in J_e}$ connected to the body part $e$ may map to multiple movement labels $\hat{y}^e_t$ of $e$ at the same time, we use the Binary Cross Entropy (BCE) loss between predicted movements $\{\hat{y}^e_t\}^{T-1}_{t=0}$ and the target movement labels $\{y^e_t\}^{T-1}_{t=0}$. This loss is minimized during the training to obtain the optimal model. During testing, the trained model of each $e \in E$ takes a sequence of 3D poses $\{\{\hat{p}^j_t\}_{j\in J_e}\}^{T-1}_{t=0}$ of all the joints connected to $e$ as input, and predicts the movement $\{\hat{y}^e_t\}^{T-1}_{t=0}$ of $e$.

\subsection{Dance Genre Recognition}
\label{sec:Genre}

Analogous to the approach in Section \ref{sec:Movement}, we train an LSTM model to take a sequence of movement labels $\{\{\hat{y}^e_t\}_{e \in E}\}^{T-1}_{t=0}$ of all the body parts $e \in E$ as input. We use the output of the last time step from the last layer as the prediction of the dance genre $\hat{g}$. For loss function, we use cross entropy between the predicted dance genre $\hat{g}$ and the target dance genre $g$. We describe the movement and dance genre recognition in detail in Algorithm \ref{alg:MovementSupplementary} 
and Algorithm \ref{alg:DanceClassifySupplementary} 
in the supplementary document.

\section{Experiments}
\label{sec:experiment}
\subsection{Data and Experiment Setting}

\begin{figure}[!ht]
\centering
\begin{tabular}{c}
\includegraphics[width=8cm,keepaspectratio]{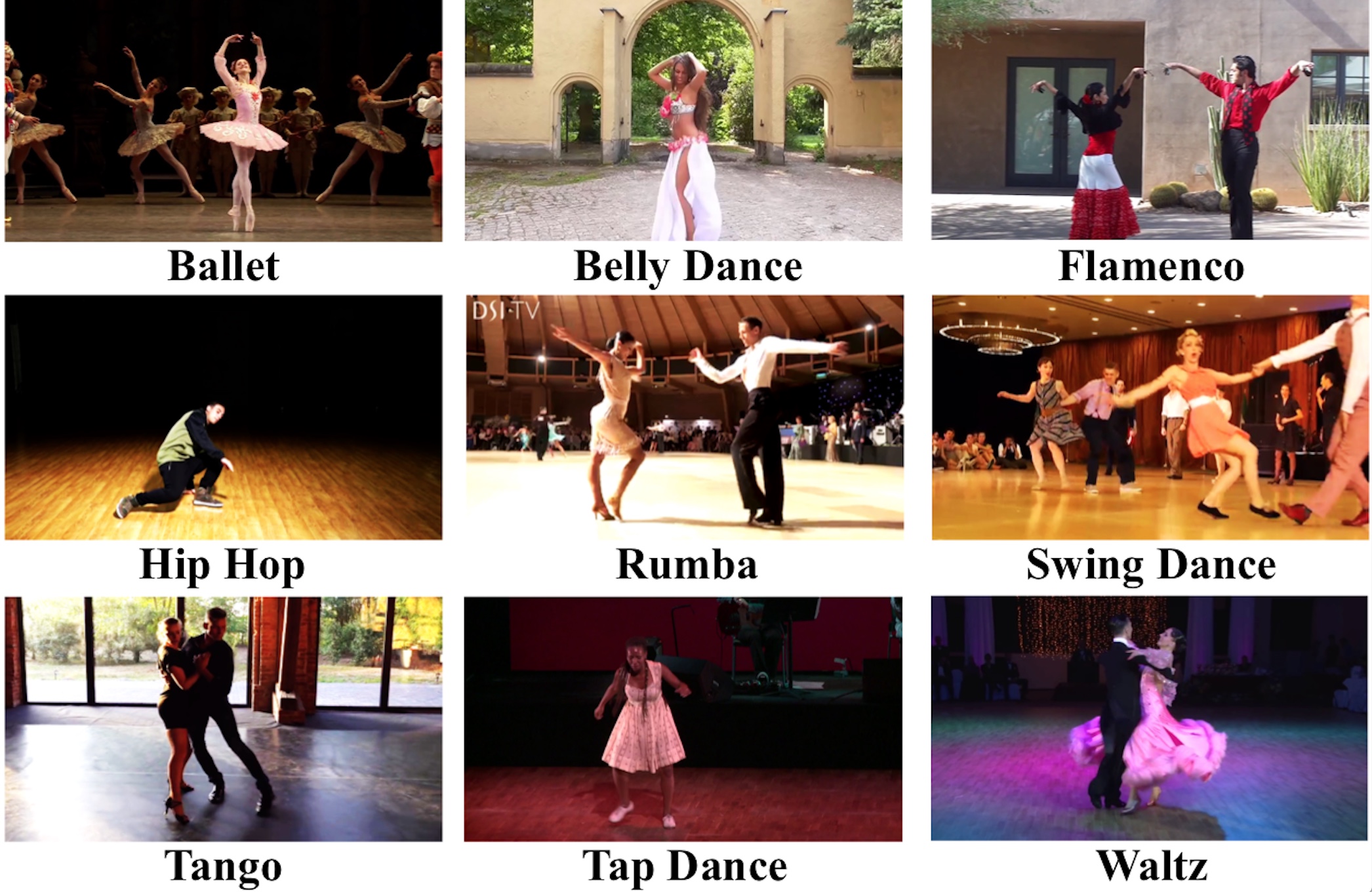}
\end{tabular}
\caption{Sample frames for 9 types of dances in the University of Illinois Dance (UID) Dataset.}
\label{fig:UID_dataset}
\end{figure}

\begin{figure}[!ht]
\centering
\begin{tabular}{c}
\includegraphics[width=8cm,keepaspectratio]{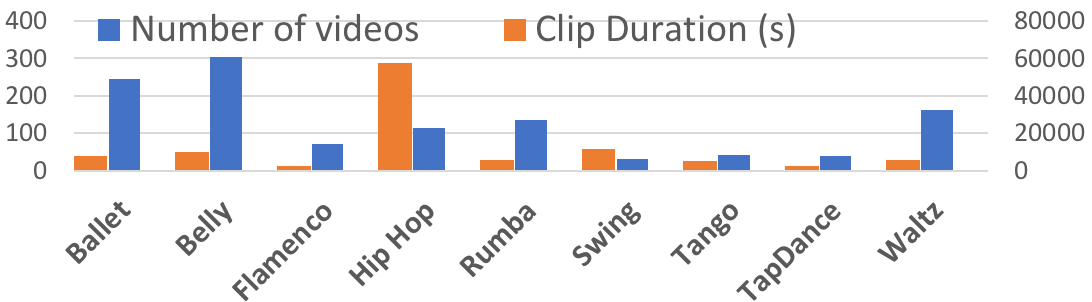}
\end{tabular}
\caption{Distribution of the numbers and durations of clips for each genre in the UID Dataset.}
\label{fig:UID_dist}
\end{figure}

\noindent\textbf{University of Illinois Dance (UID) Dataset.}
One major challenge for dance recognition lies in the lack of training data. We have curated \textit{UID video dataset} containing 9 types of dances (Ballet, Belly dance, Flamenco, Hip Hop, Rumba, Swing dance, Tango, Tap dance and Waltz) with details listed in Table \ref{tab:UID_dataset}. Figure \ref{fig:UID_dataset} and \ref{fig:UID_dist} show sample frames and information about in our dataset for each dance genre. The videos contain situations of varying difficulty, from simple ones such as tutorial videos with clean background, to hard videos, having interacting dancers, noisy background and varying lights.

\begin{table}[!ht]
\small
\centering
\begin{tabular}{l|l|l|l}
	\toprule
	\textbf{Dance Genres} & 9 &
	\textbf{Total Duration} & 108,089s \\
	\hline
	\textbf{Total \# of Clips} & 1143 & \textbf{Total \# of Frames} & 2,788,157 \\
	\hline
	\textbf{Min clip length} & 4s &
	\textbf{Min \# of clips / class} & 30 \\
	\hline
	\textbf{Max clip length} & 824s & \textbf{Max \# of clips / class} & 304 \\
	\bottomrule
\end{tabular}
\caption{Summary of the characteristics of the UID dataset.}
\label{tab:UID_dataset}
\end{table}

\begin{table}[!ht]
\small
\centering
\begin{tabular}{llll}
	\toprule
	\parbox{0.7cm}{\textbf{Method}} & \parbox{0.8cm}{\textbf{Supervision}} & \parbox{0.7cm}{\textbf{Extra Data}} & \parbox{1.2cm}{\textbf{MPJPE (mm)$(\downarrow)$}} \\
	\midrule
    Martinez \cite{Martinez2017ICCV} {\scriptsize ICCV'17} & Supervised & - & 110.0 \\
    Wandt \cite{Wandt2019RepNet} {\scriptsize CVPR'19} & Supervised & - & 323.7 \\
	Pavllo \cite{Pavllo2019CVPR} {\scriptsize CVPR'19} & Supervised & - & \underline{77.6} \\
	\midrule
	Pavllo \cite{Pavllo2019CVPR} {\scriptsize CVPR'19$(\star)$} & Semi-Sup. & No & 446.1 \\
	Ours & Semi-Sup. & No & \textbf{73.7} \\
	\midrule
	Zhou \cite{Zhou2017ICCV} {\scriptsize ICCV'17} & Weakly-Sup. & Yes & 93.1 \\
	Kocabas \cite{kocabas2019epipolar} {\scriptsize CVPR'19} & Self-Sup. & {\scriptsize Multiview} & 87.4 \\
	Ours & Unsupervised & No & 246.4 \\
	\bottomrule
\end{tabular}
\caption{Comparison of 3D pose estimation results using Protocol 1: Mean Per-Joint Position Error (MPJPE) on AIST Dance Video Dataset \cite{aist}. $(\star)$ uses ground truth 2D poses. Methods using different supervision level are divided by horiontal line. Our proposed method (semi-supervised) achieves the lowest error against the fully supervised methods. Moreover, our unsupervised pose estimation method can achieve the same level of performance as the state-of-the-art supervised/semi-supervised methods.
}
\label{tab:3Dpose_result_aist}
\end{table}

\begin{table}[!ht]
\small
\centering
\begin{tabular}{llll}
	\toprule
	\parbox{0.8cm}{\textbf{Method}} & \parbox{0.8cm}{\textbf{Supervision}} & \parbox{0.7cm}{\textbf{Extra Data}} & \parbox{1.2cm}{\textbf{MPJPE (mm)$(\downarrow)$}} \\
	\midrule
    Martinez \cite{Martinez2017ICCV} {\scriptsize ICCV'17} & Supervised & - & 87.3 \\
    Zanfir \cite{Zanfir2018CVPR} {\scriptsize CVPR'18} & Supervised & - & 69.0 \\
    Wandt \cite{Wandt2019RepNet} {\scriptsize CVPR'19} & Supervised & - & 89.9 \\
	Pavllo \cite{Pavllo2019CVPR} {\scriptsize CVPR'19} & Supervised & - & \textbf{46.8} \\
    Mehta \cite{XNect2020SIGGRAPH} {\tiny SIGGRAPH'20} & Supervised & - & 63.6 \\
	\midrule
	Pavllo \cite{Pavllo2019CVPR} {\scriptsize CVPR'19$(\star)$}  & Semi-Sup. & No & 51.6 \\
	Ours & Semi-Sup. & No & \underline{47.3} \\
	\midrule
	Zhou \cite{Zhou2017ICCV} {\scriptsize ICCV'17} & Weakly-Sup. & Yes & 64.9 \\
	Rhodin \cite{rhodin2018ECCV} {\scriptsize ECCV'18} & Unsupervised & {\scriptsize Multiview} & 98.2 \\
	Kocabas \cite{kocabas2019epipolar} {\scriptsize CVPR'19} & Self-Sup. & {\scriptsize Multiview} & 60.6 \\
	Chen \cite{Chen2019cvpr} {\scriptsize CVPR'19} & Unsupervised & Yes & 68.0 \\
	Kundu \cite{kundu2020unsup} {\scriptsize ECCV'20} & Unsupervised & Yes & 67.9 \\
	Ours & Unsupervised & No & 82.1 \\
	\bottomrule
\end{tabular}
\caption{Comparison of 3D pose estimation results using Protocol 1: Mean Per-Joint Position Error (MPJPE) on Human3.6M Dataset \cite{h36m} evaluated
on S9 and S11. $(\star)$ uses ground truth 2D poses. Based on the method's supervision level, five labelled subjects (S1, S5, S6, S7, S8) are used to train the supervised methods, four labelled subjects (S5, S6, S7, S8) and one unlabelled subject (S1) are used to train the semi-supervised methods, and five  unlabelled subjects (S1, S5, S6, S7, S8) for the rest methods (e.g., unsupervised).
Our proposed method (semi-supervised) achieves the second lowest error against the fully supervised methods. Without the need of additional 2D/3D data, our unsupervised pose estimation method can achieve the same level of performance as the state-of-the-art methods.
}
\label{tab:3Dpose_result_human36}
\end{table}

\begin{figure*}[!ht]
\centering
\scalebox{0.85}{
\setlength{\tabcolsep}{1pt}
\begin{tabular}{ccccccccc}
\multirow{2}{*}{\rotatebox[origin=c]{90}{\small{AIST++ Dataset}}} & \ctab{\includegraphics[height=1.3cm,trim=15cm 5cm 15cm 10cm ,clip]{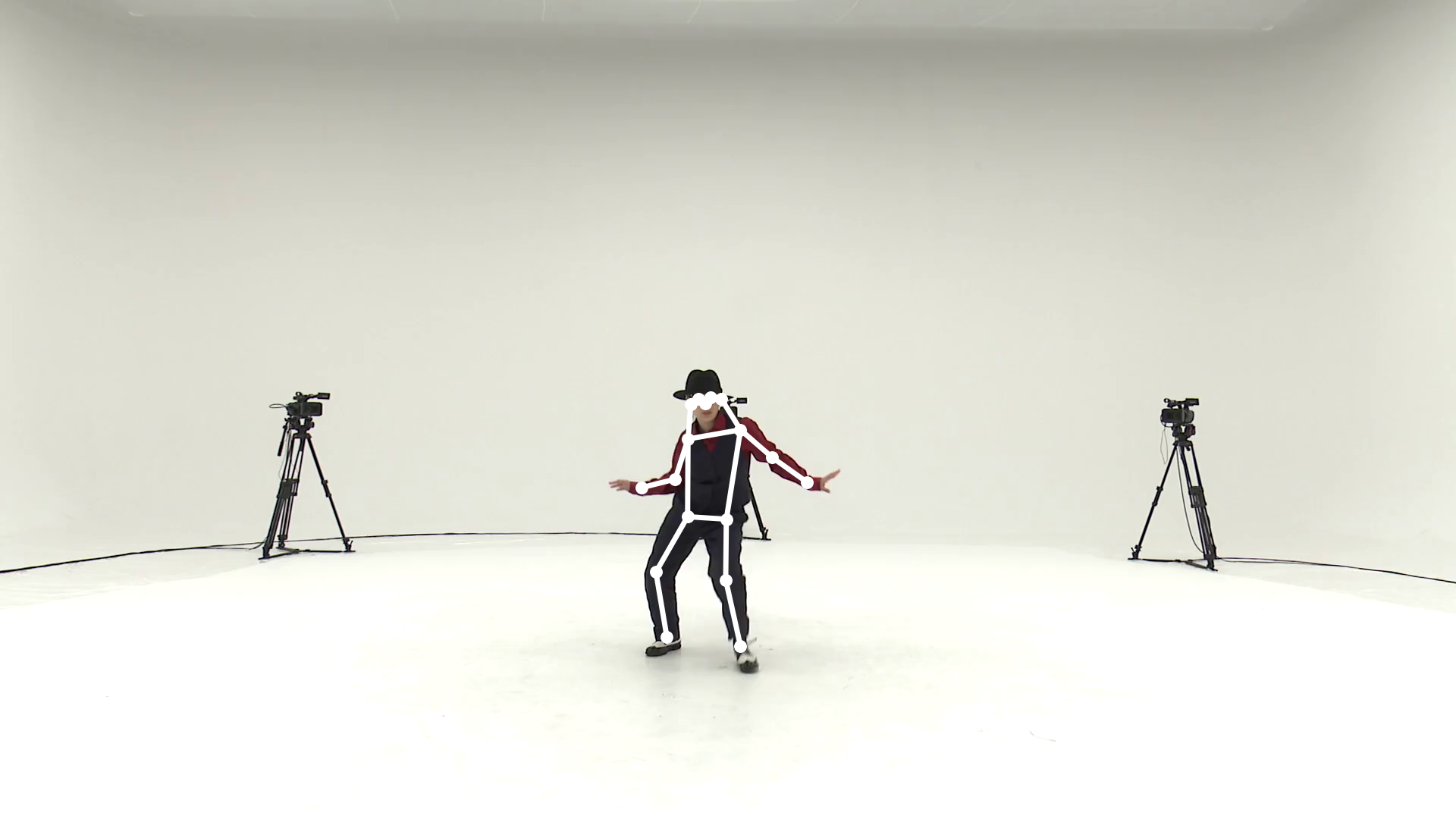}} &
\ctab{\includegraphics[height=1.3cm,trim=15cm 5cm 15cm 10cm ,clip]{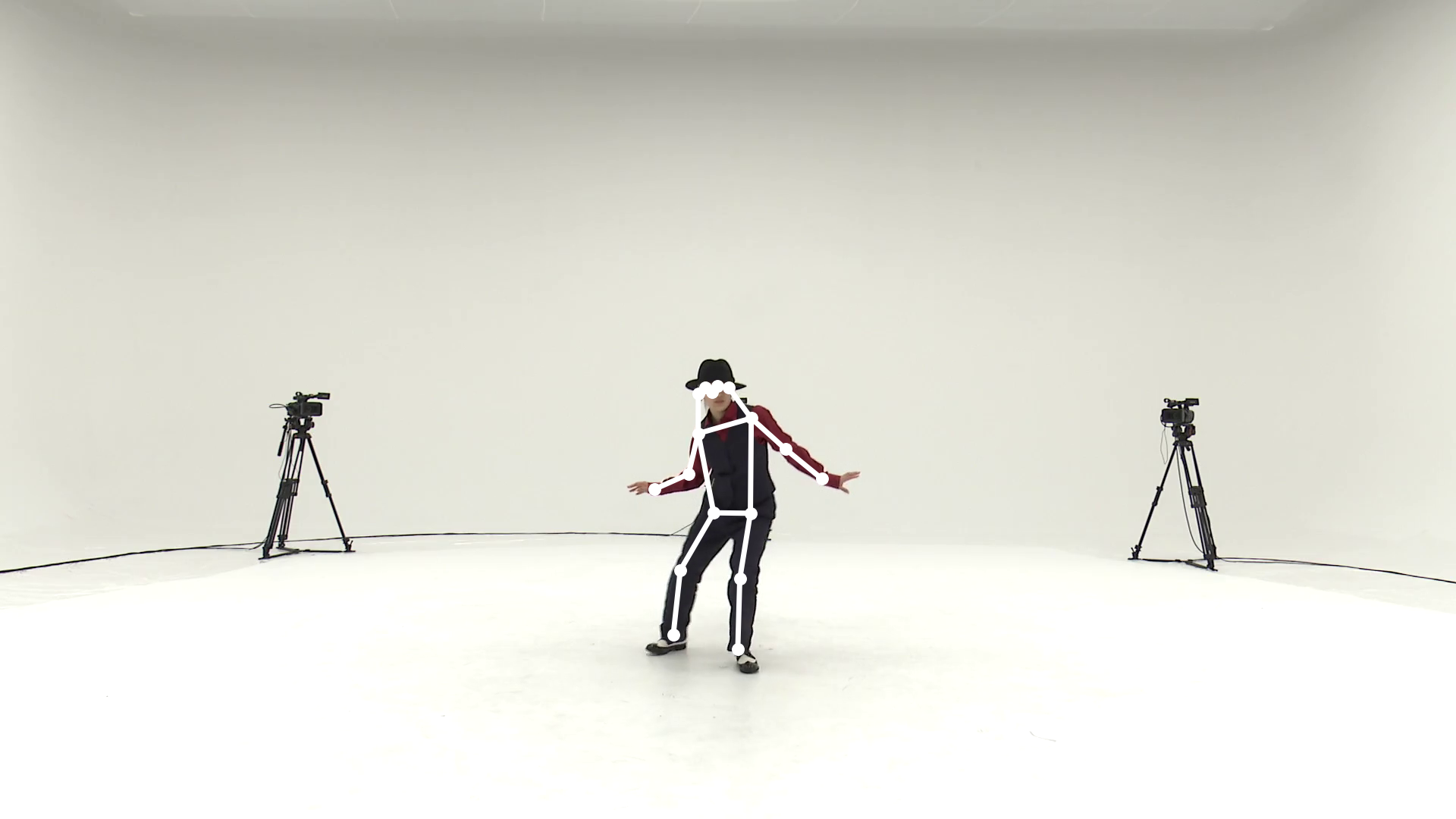}} &
\ctab{\includegraphics[height=1.3cm,trim=15cm 5cm 15cm 10cm ,clip]{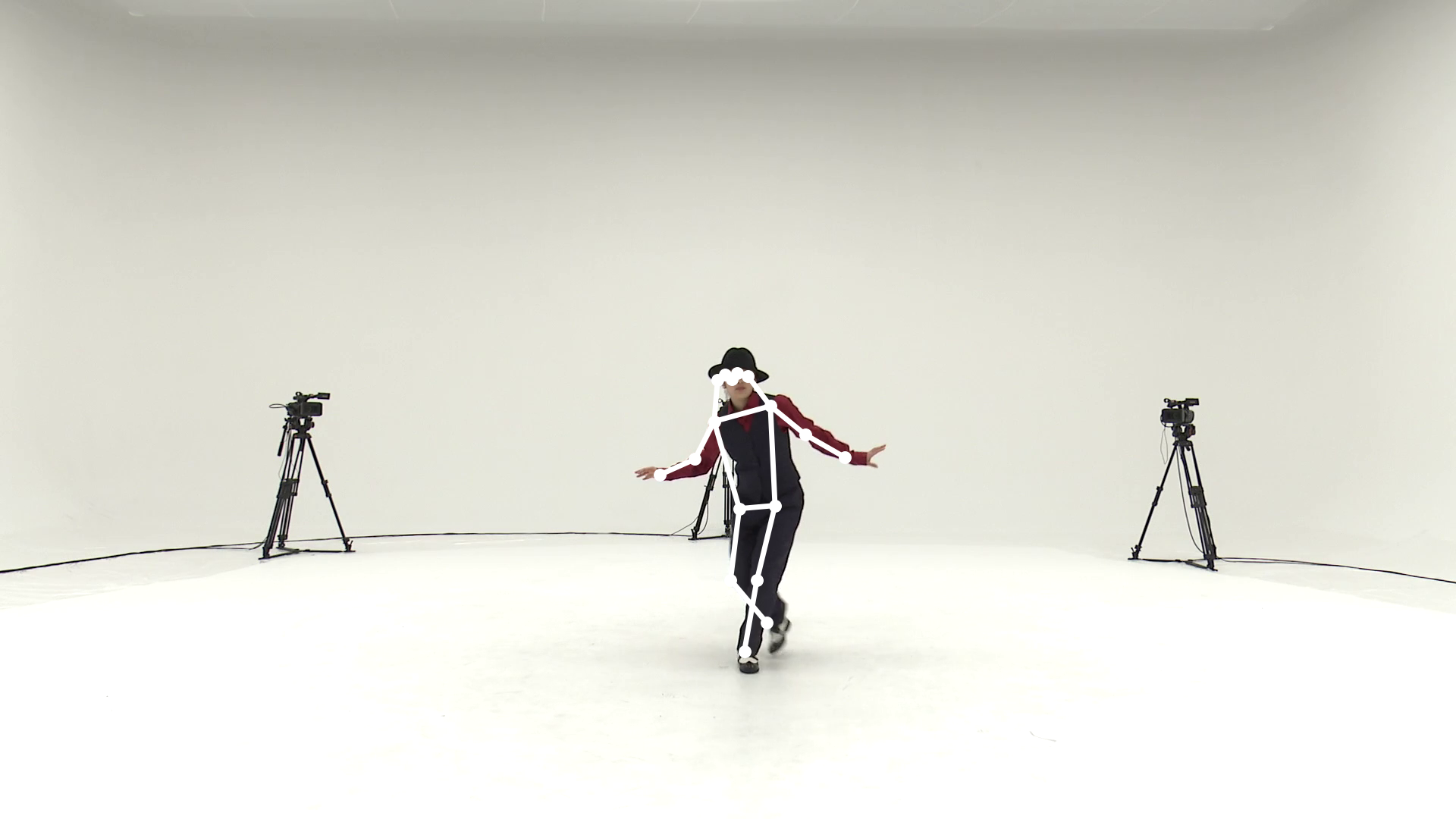}} &
\ctab{\includegraphics[height=1.3cm,trim=15cm 5cm 15cm 10cm ,clip]{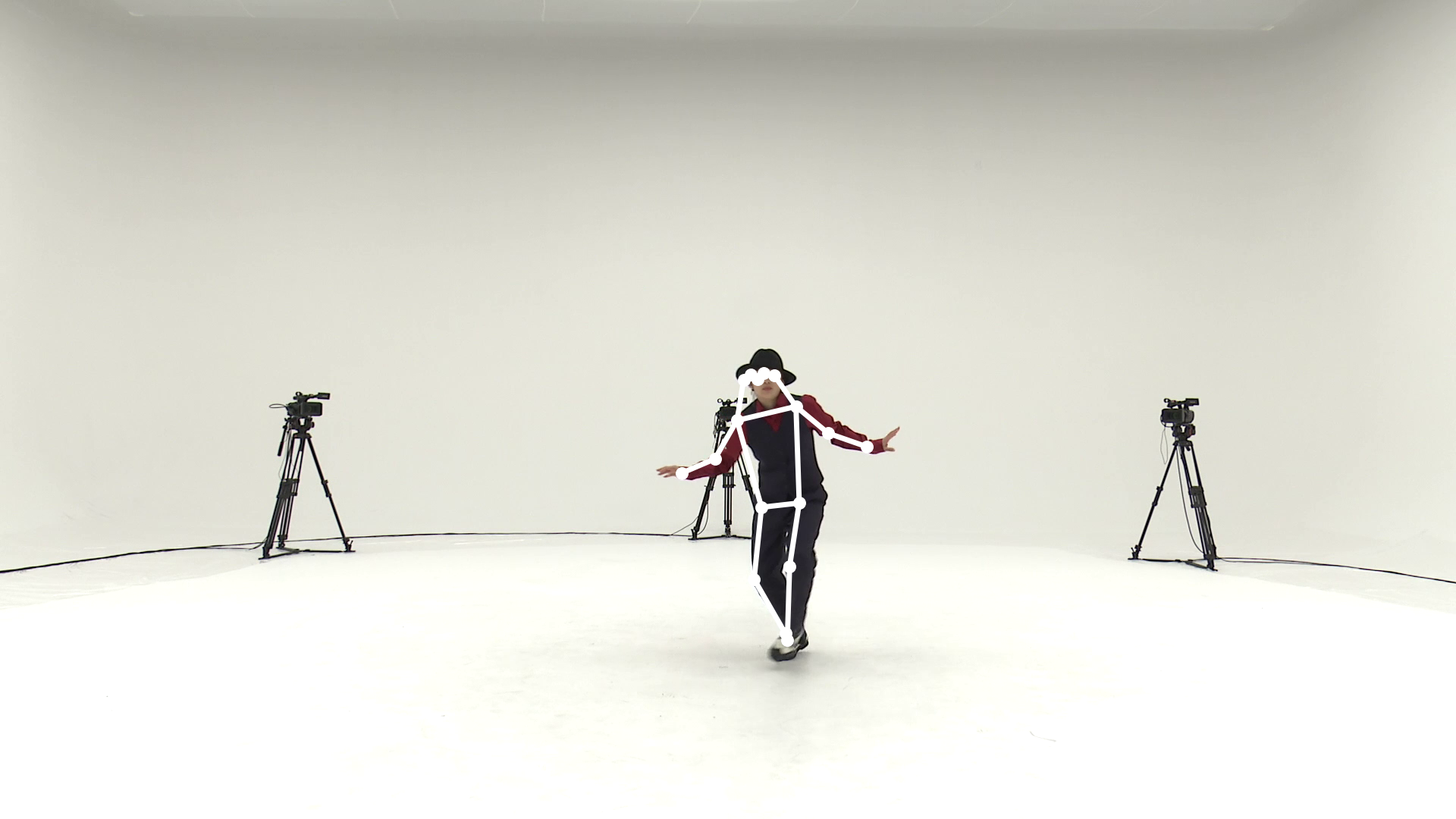}} &
\ctab{\includegraphics[height=1.3cm,trim=15cm 3cm 15cm 8cm ,clip]{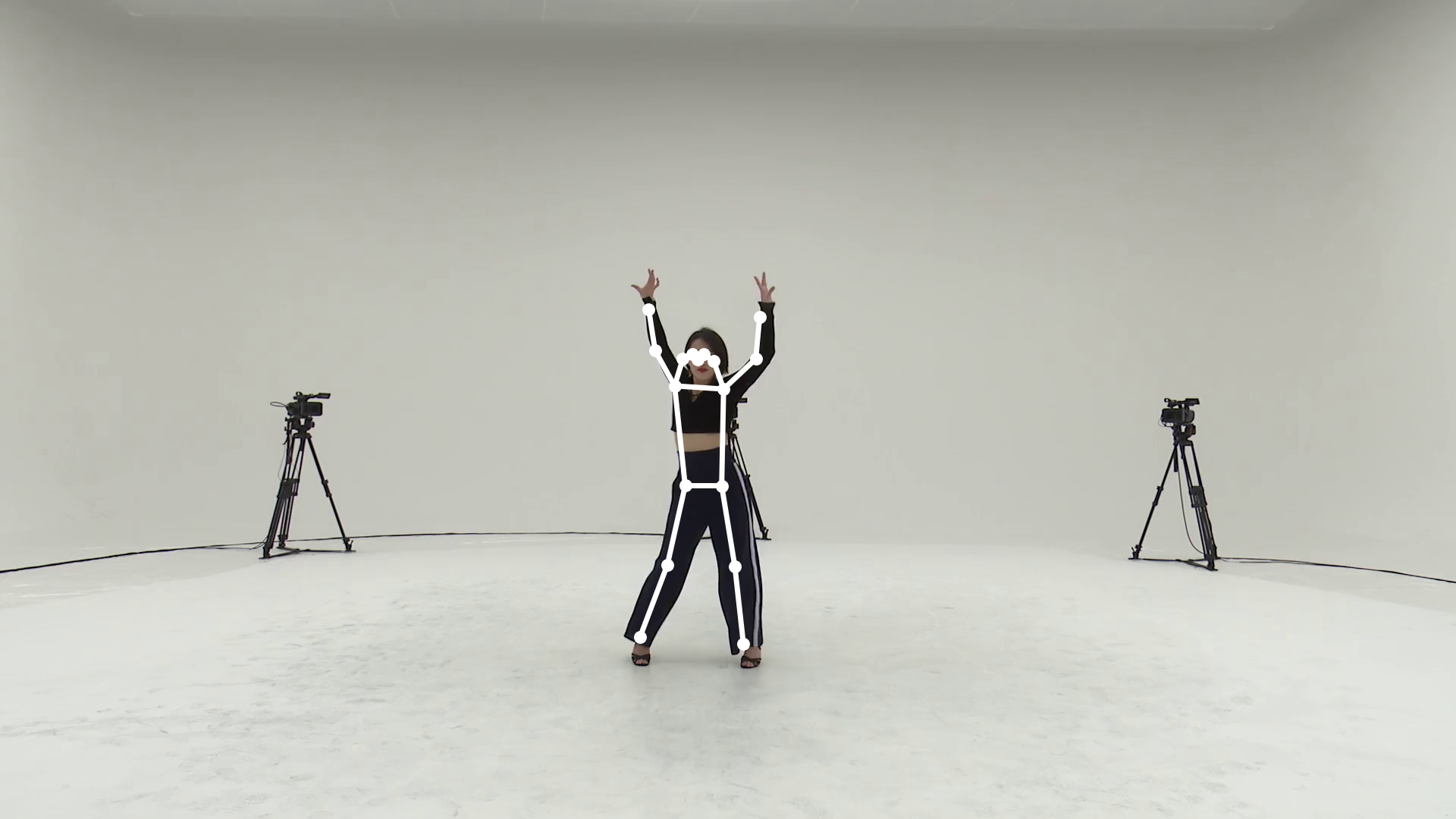}} &
\ctab{\includegraphics[height=1.3cm,trim=15cm 3cm 15cm 8cm ,clip]{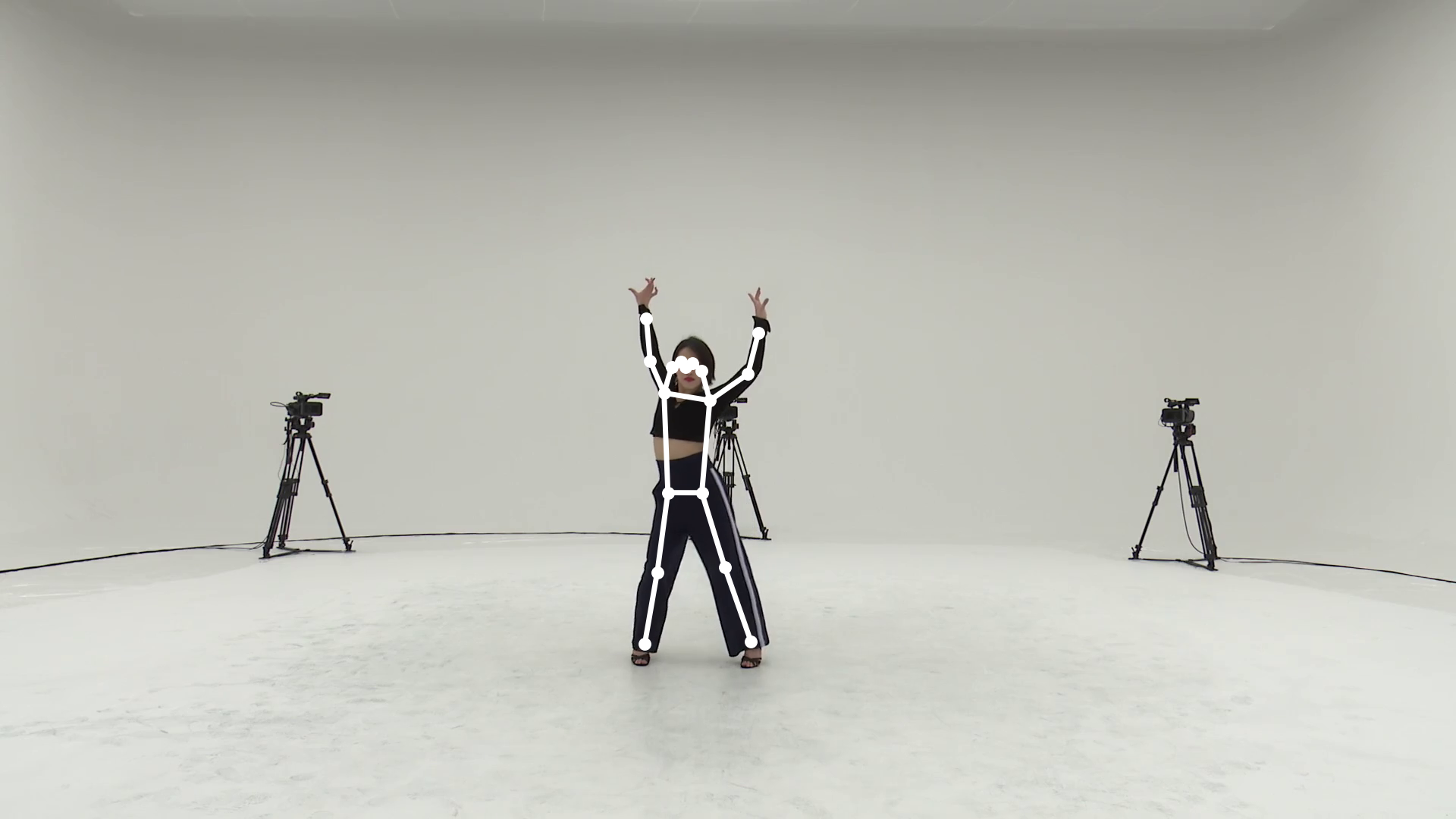}} &
\ctab{\includegraphics[height=1.3cm,trim=15cm 3cm 15cm 8cm ,clip]{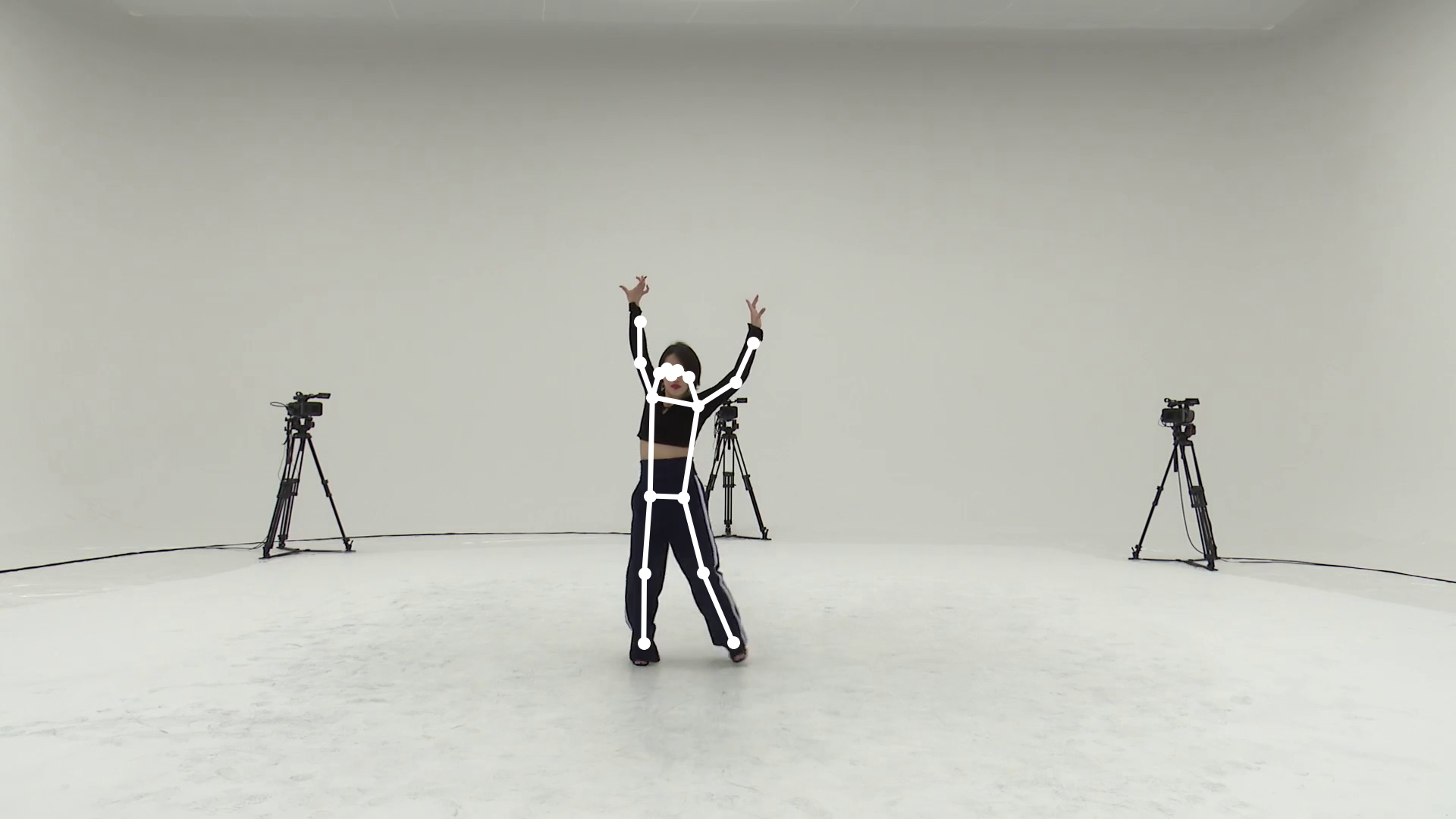}} &
\ctab{\includegraphics[height=1.3cm,trim=15cm 3cm 15cm 8cm ,clip]{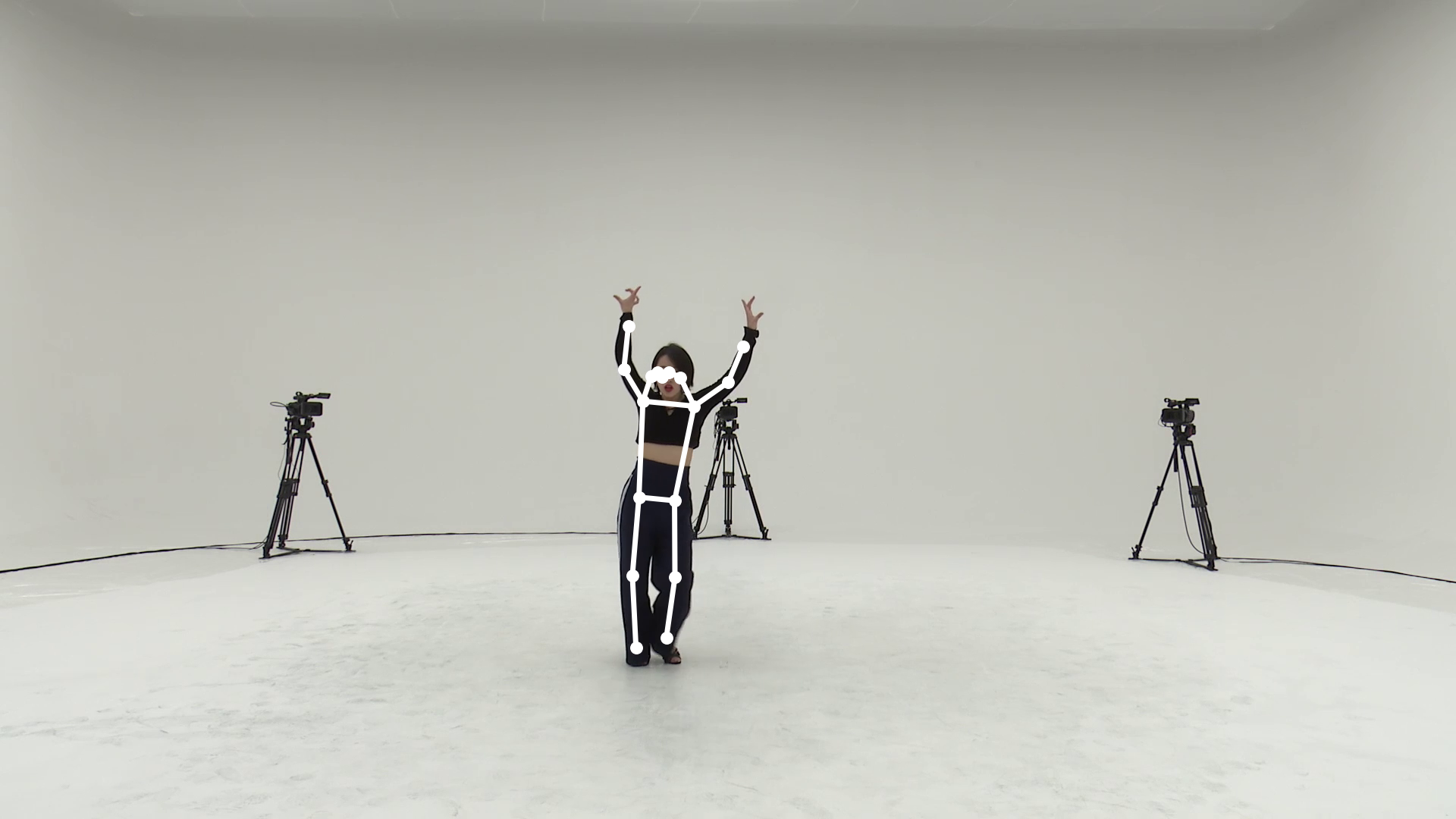}}\\
& \ctab{\includegraphics[height=1.3cm,trim=3cm 3cm 3cm 3cm ,clip]{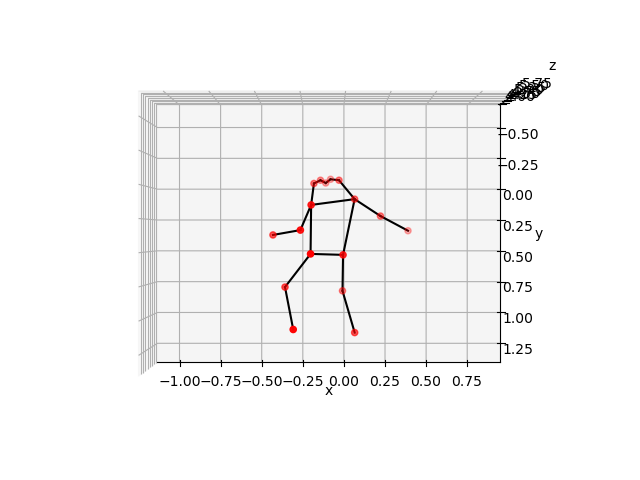}} &
\ctab{\includegraphics[height=1.3cm,trim=3cm 3cm 3cm 3cm ,clip]{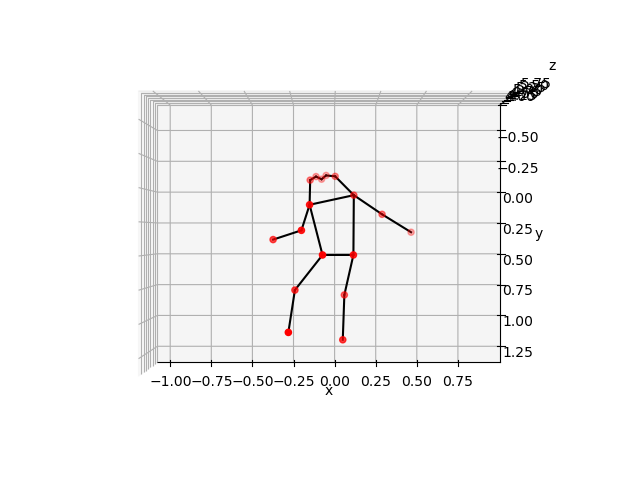}} &
\ctab{\includegraphics[height=1.3cm,trim=3cm 3cm 3cm 3cm ,clip]{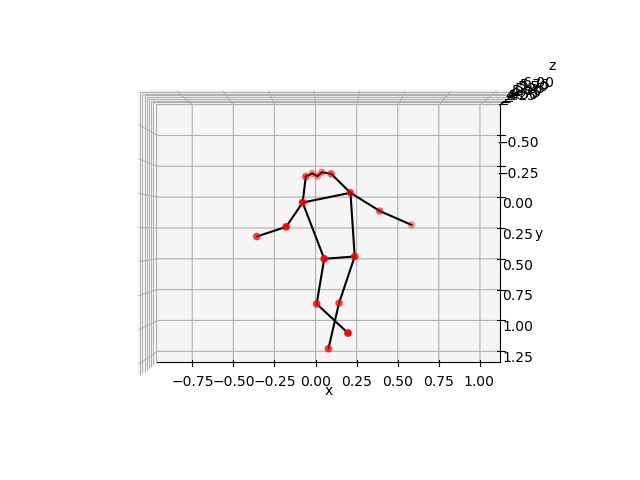}} &
\ctab{\includegraphics[height=1.3cm,trim=3cm 3cm 3cm 3cm ,clip]{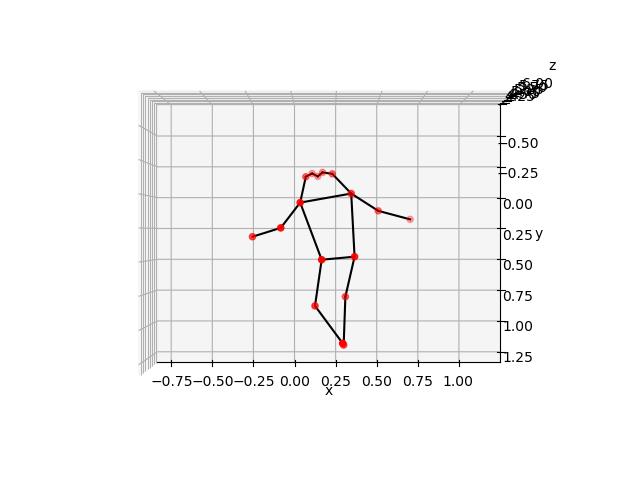}} &
\ctab{\includegraphics[height=1.3cm,trim=3cm 3cm 3cm 3cm ,clip]{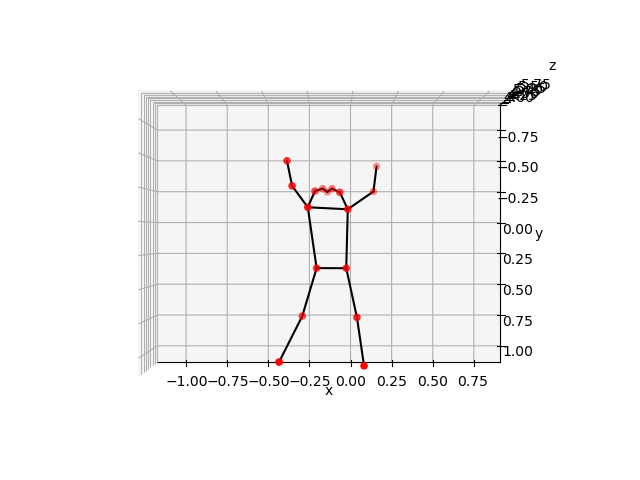}} &
\ctab{\includegraphics[height=1.3cm,trim=3cm 3cm 3cm 3cm ,clip]{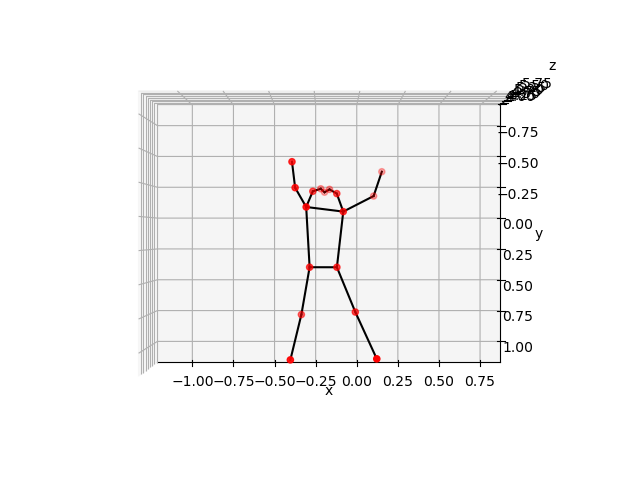}} &
\ctab{\includegraphics[height=1.3cm,trim=3cm 3cm 3cm 3cm ,clip]{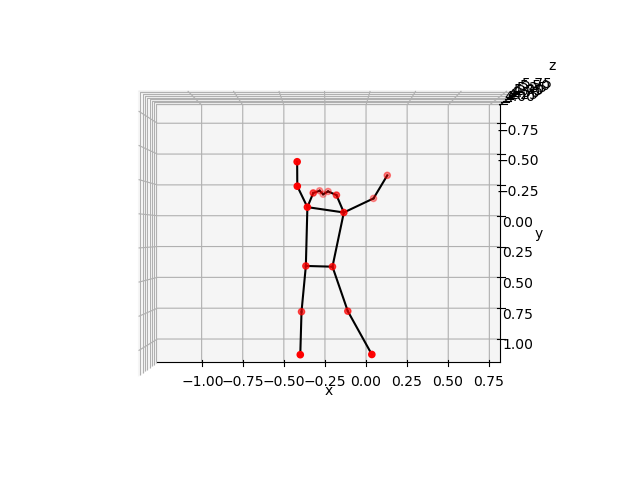}} &
\ctab{\includegraphics[height=1.3cm,trim=3cm 3cm 3cm 3cm ,clip]{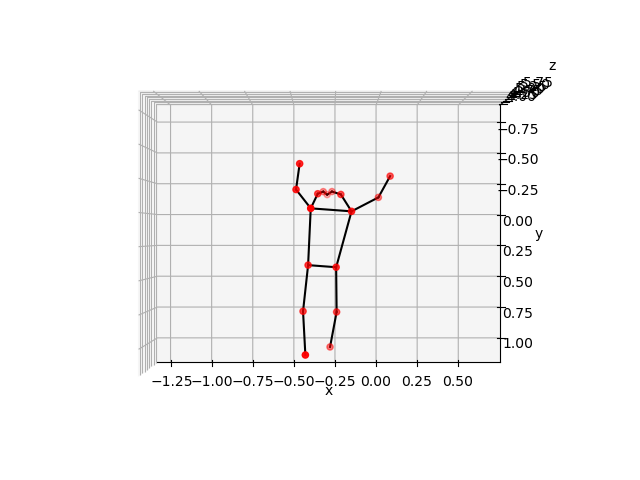}}\\
\multirow{2}{*}{\rotatebox[origin=c]{90}{\small{UID Dataset}}} & \ctab{\includegraphics[height=1.3cm,trim=1cm 0cm 1cm 0cm ,clip]{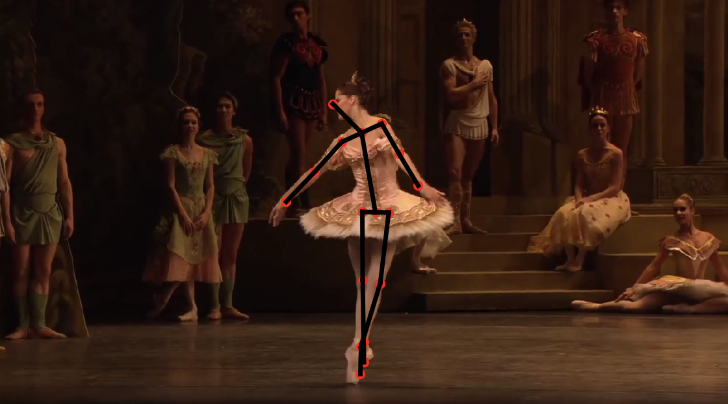}} &
\ctab{\includegraphics[height=1.3cm,trim=1cm 0cm 1cm 0cm,clip]{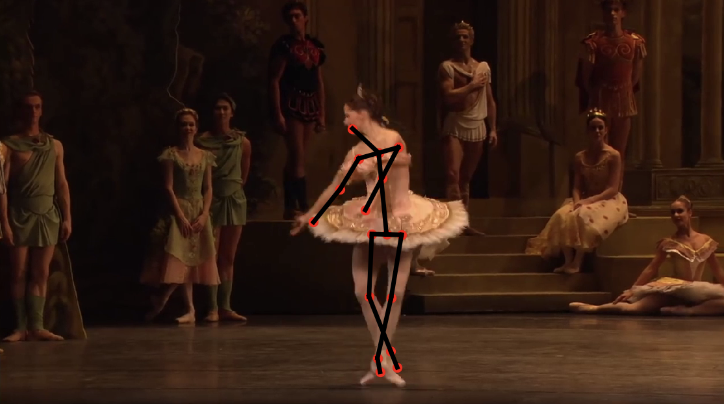}} &
\ctab{\includegraphics[height=1.3cm,trim=1cm 0cm 1cm 0cm,clip]{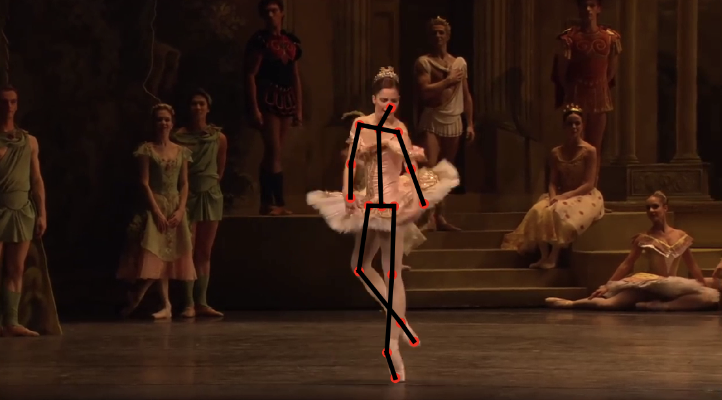}} &
\ctab{\includegraphics[height=1.3cm,trim=1cm 0cm 1cm 0cm,clip]{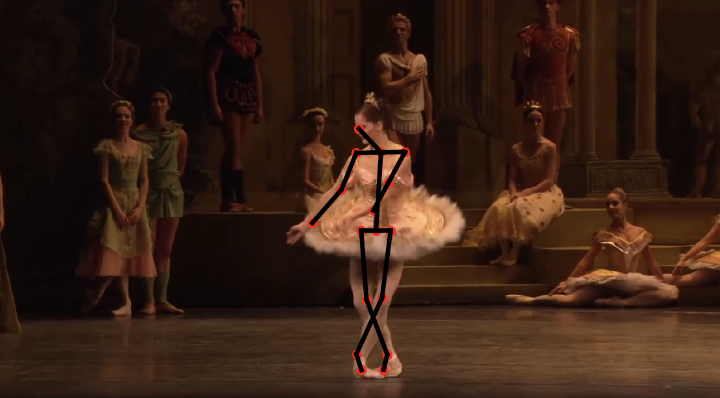}} &
\ctab{\includegraphics[height=1.3cm,trim=1cm 0cm 1cm 0cm,clip]{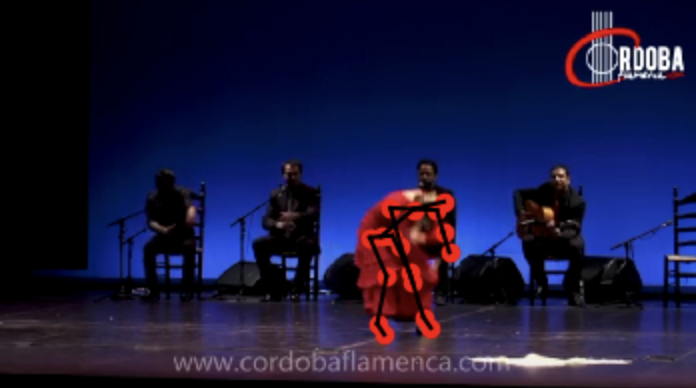}} &
\ctab{\includegraphics[height=1.3cm,trim=1cm 0cm 1cm 0cm,clip]{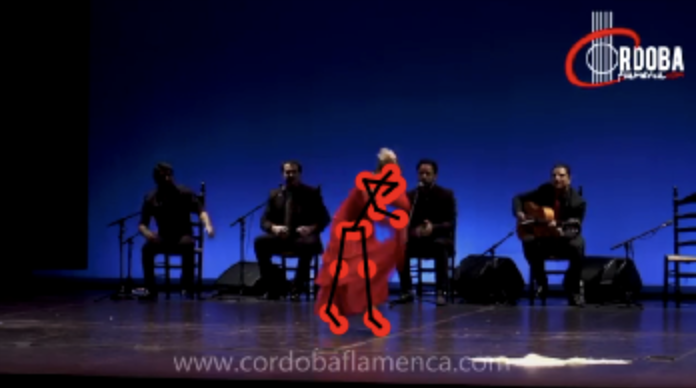}} &
\ctab{\includegraphics[height=1.3cm,trim=1cm 0cm 1cm 0cm,clip]{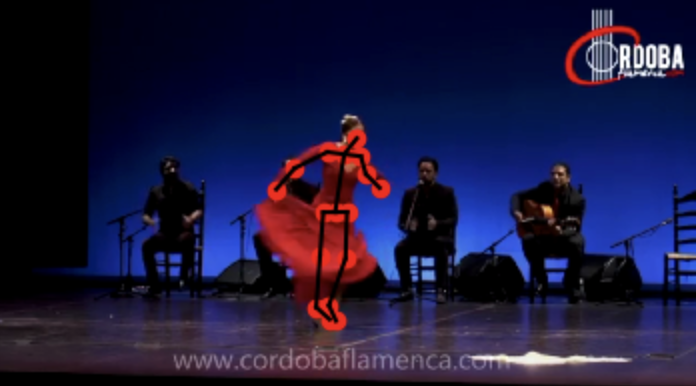}} &
\ctab{\includegraphics[height=1.3cm,trim=1cm 0cm 1cm 0cm,clip]{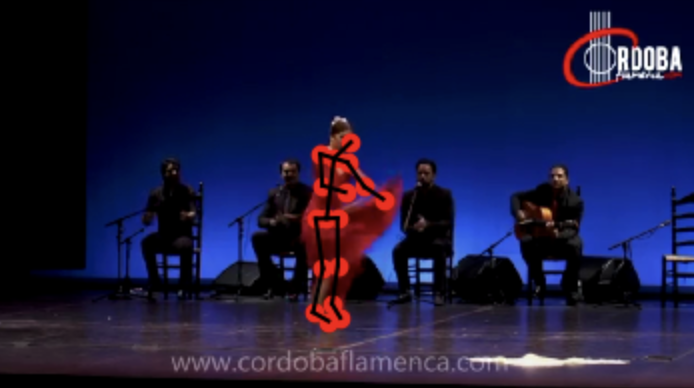}} \\
& \ctab{\includegraphics[height=1.3cm,trim=1cm 1cm 1cm 1cm ,clip]{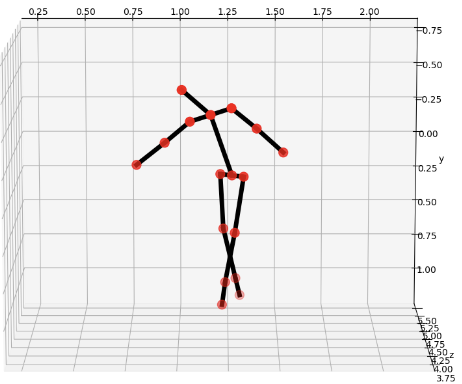}} &
\ctab{\includegraphics[height=1.3cm,trim=1cm 1cm 1cm 1cm,clip]{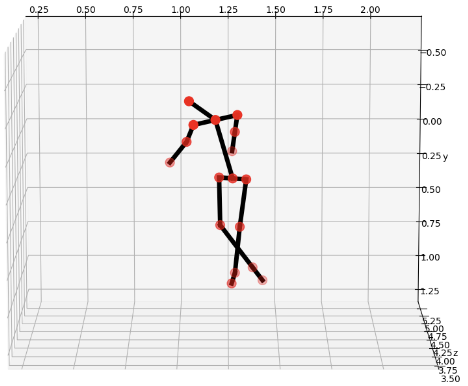}} &
\ctab{\includegraphics[height=1.3cm,trim=1cm 1cm 1cm 1cm,clip]{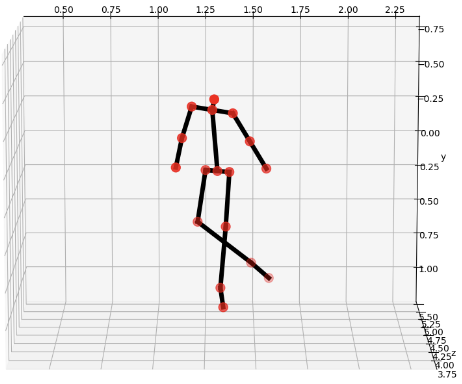}} &
\ctab{\includegraphics[height=1.3cm,trim=1cm 1cm 1cm 1cm,clip]{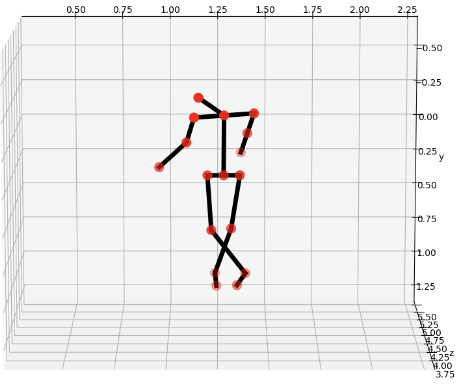}} &
\ctab{\includegraphics[height=1.3cm,trim=1cm 1cm 1cm 1cm,clip]{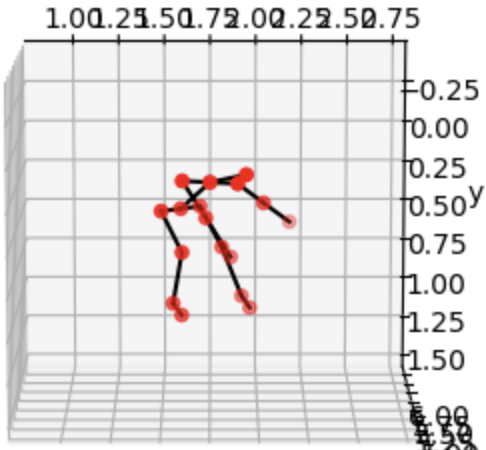}} &
\ctab{\includegraphics[height=1.3cm,trim=1cm 1cm 1cm 1cm,clip]{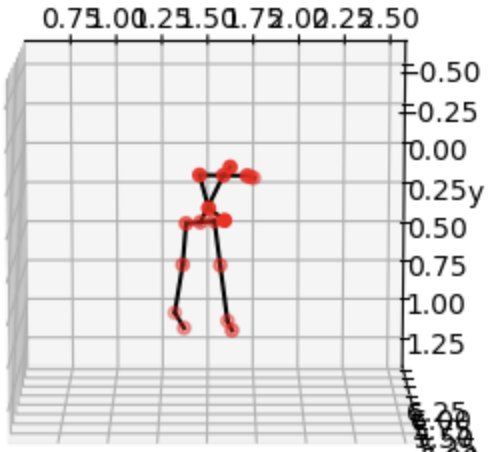}} &
\ctab{\includegraphics[height=1.3cm,trim=1cm 1cm 1cm 1cm,clip]{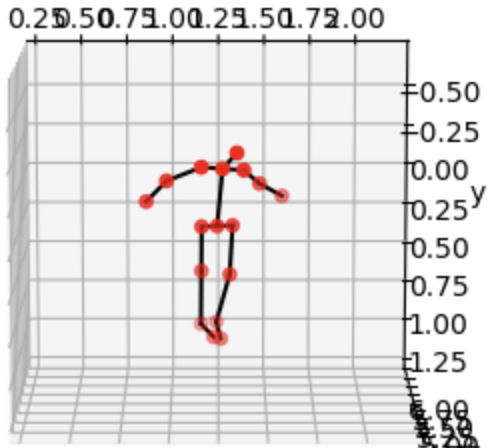}} &
\ctab{\includegraphics[height=1.3cm,trim=1cm 1cm 1cm 1cm,clip]{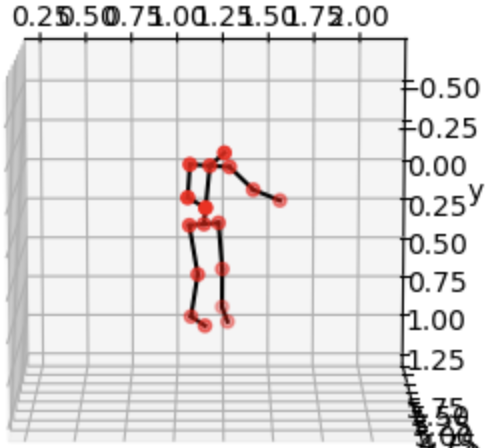}} \\
& Frame 0 & Frame 10 & Frame 20 & Frame 30 & Frame 0 & Frame 10 & Frame 20 & Frame 30 \\
\end{tabular}}
\caption{Visualization results on sample videos from the AIST++ dataset \cite{li2021learn} and our proposed University of Illinois Dance (UID) dataset. The top row shows the reconstructed 2D poses from the estimated 3D poses and the bottom row shows the estimated 3D poses.}
\label{fig:qualitative}
\end{figure*}

\begin{table*}[!ht]
\small
\centering
\begin{tabular}{l|l|lllllllllll}
	\toprule
	\multirow{2}{*}{\parbox{2.2cm}{\raggedright Input to the \mbox{Movement Recog.}}} & \multicolumn{12}{c}{F-score} \\
	\cline{2-13}
	& \textit{Averaged} & Head & lshoulder & rshoulder & larm & rarm & Hips & Torso & lleg & rleg & lfoot & rfoot \\
	\midrule
    2D Pose & 0.93 & \textbf{0.95} & \textbf{0.96} & \textbf{0.96} & 0.89 & 0.91 & 0.81 & 0.96 & 0.94 & 0.85 & \textbf{1.00} & \textbf{1.00}
 \\
    3D Pose & \textbf{0.97} & 0.93 & \textbf{0.96} & \textbf{0.96} & \textbf{0.94} & \textbf{0.93} & \textbf{1.00} & \textbf{0.98} & \textbf{0.95} & \textbf{0.98} & 0.99 & \textbf{1.00}
 \\
	\bottomrule
\end{tabular}
\caption{F-scores for body part movements recognition from estimated 2D poses (Sec \ref{sec:2Dpose}) and estimated 3D poses (Sec \ref{sec:3Dpose}) as inputs. Recognition improves as a result of using our estimated 3D poses. Note that the performances for several parts are comparable with existing results. This is because the dancers are at a large distance, diminishing the extra power offered by the 3D information. This situation changes in Table \ref{tab:genre_recog}.
}
\label{tab:move_recog}
\end{table*}

\begin{table}[!ht]
\small
\centering
\begin{tabular}{l|l}
	\toprule
	Input to Dance Genre Recognition & F-score \\
	\midrule
    2D Pose & 0.44 \\
    3D Pose & 0.47 \\
    Movements (2D Pose as input) & 0.50 \\
    Movements (3D Pose as input) & 0.55 \\
    2D Pose + Movements (2D Pose as input) & \textbf{0.73} \\
    3D Pose + Movements (3D Pose as input) & \textbf{0.86} \\
	\bottomrule
\end{tabular}
\caption{Ablation study using different components as inputs. The 3D pose, in general, provides higher accuracy for genre recognition than 2D pose. Combination of the two, 2D and 3D level estimates, achieves better performance than either alone.
}
\label{tab:genre_recog}
\end{table}

\noindent\textbf{Evaluation Protocols.} 
%
%
we use the widely used mean per-joint position error (MPJPE) in millimeters to calculate the mean Euclidean distance between the predicted 3D poses $\{\hat{P}_t\}_{t=0}^{T-1}$ and the target 3D poses $\{P_t\}_{t=0}^{T-1}$.
We use F-score to measure the accuracy of our movement and dance recognition approaches on our UID dataset.

\noindent\textbf{Experiment Setting.} We evaluate our unsupervised 3D pose estimation approach on both the UID video dataset and AIST++ dance dataset \cite{aist}. The AIST++ Dataset contains 1,408 multi-view dance sequences from 10 dance genres with hundreds of choreographies, provides 3D human keypoint annotations and camera parameters for 10.1M images, and covers 30 different subjects in 9 views. We did our experiments with a subset of AIST++, containing 200 videos (~0.4M frames). 30\% of the videos with ground-truth 3D poses are used as labeled data to train the supervised methods \cite{Wandt2019RepNet, Martinez2017ICCV, Pavllo2019CVPR} and semi-supervised methods (\cite{Pavllo2019CVPR} and our method). 10\% of the videos are used for testing. The remaining video samples are used as unlabeled data for training the semi-supervised methods. 

For consistency with other work \cite{Wandt2019RepNet, Martinez2017ICCV, Pavllo2019CVPR}, we train and evaluate on $3D$ poses in camera space. In the 3D Pose Initialization component, we use Adam \cite{kingma2017adam} optimizer to optimize the estimated 3D poses in Algorithm \ref{alg:3D_to_2D} for 50 epochs. The temporal window size $\Delta=3$ and the number of seeds $K=2$. After obtaining the best initial 3D poses and camera projection parameters (focal lengths and principal points), we use \cite{Pavllo2019CVPR} as the baseline to train the 3D pose estimation network for 200 epochs.

\subsection{3D Poses}

Figure \ref{fig:qualitative} shows qualitative results of our 3D pose method on both the UID dataset and the AIST++ dataset \cite{aist}. The 2D poses (top row) reconstructed from the estimated 3D poses align well with the dancer's movement. The estimated 3D poses well match the known human skeletal structure and are smooth between frames.
To quantitatively evaluate our method, we train our model and three state-of-the-art methods \cite{Wandt2019RepNet, Martinez2017ICCV, Pavllo2019CVPR} on the AIST++ dataset and calculate the mean per-joint position errors (MPJPE). We also evaluated our model on the Human 3.6M dataset \cite{h36m}. Table \ref{tab:3Dpose_result_aist} and Table \ref{tab:3Dpose_result_human36} shows that our unsupervised pose estimation method is comparable with the supervised methods. Moreover, our semi-supervised version achieves the best and second best performance on the AIST++ dataset \mbox{\cite{aist}} and 3.6M dataset \mbox{\cite{h36m}}, respectively.

\subsection{Movement and Dance Genre Recognition}
Recognition results for body part movements and dance genre recognition on the UID dataset are given in Tables \ref{tab:move_recog} and \ref{tab:genre_recog}. {We use the 3D poses estimated using our unsupervised method as the input for recognition since our UID collects videos in the wild and hence does not provide ground-truth 3D annotations for training the proposed semi-supervised version. The movements of different body parts can help with dance understanding from the viewpoint of dance experts.}

\section{Conclusions and Future Work}
In conclusion,we have presented an approach to dance videos understanding that follows a hierarchical representation used by experts to describe dances. We have presented an approach to extract the primitives occurring at each level of the representation, from raw videos, to 3D pose, to movements, to dance genre. We have presented the challenges we have encountered and how we have addressed them using new constraints and algorithms. 
{
Note that the training in our current dance video recognition framework is not fully unsupervised. We plan to develop a fully unsupervised pipeline that could be jointly trained for pose estimation and genre recognition.
}
In addition, we plan to synthesize dances using the representations we have extracted. 
We also plan to use the judgments of expert viewers on the quality of the synthesized dance videos as qualitative metrics of the representations extracted by our algorithms.

\clearpage
{\small
\bibliographystyle{unsrt}
\bibliography{iccv2021}
}

\onecolumn
\begin{appendices}
\section{UID Dataset}
The proposed \textit{UID video dataset} can be found at \url{https://drive.google.com/drive/folders/1-SdWYxIorbhQzi9Bp_HpJf25_ieMjoh5?usp=sharing}.
The dataset folder contains 9 sub-folders, each having $\sim$30 to $\sim$300 videos, showing one of the following 9 dance types: Ballet, Belly dance, Flamenco, Hip Hop, Rumba, Swing dance, Tango, Tap  dance and Waltz.

\section{Demo Videos}
In Figure \ref{fig:qualitative} in Section \ref{sec:experiment} of the paper, we have shown estimated 3D poses by drawing skeletal figures in the estimated poses and overlaying them on the corresponding images of the dancers. These single frame overlays help us verify the placement of the skeleton within the body parts in only four frames in a video (containing $\sim$100 to $\sim$800 frames).
However, the contributions of our paper also include enforcement of temporal smoothness constraints, and estimation of complex 3D poses. Here we therefore include videos that show the overlays of the skeletons in all frames of the videos. Viewing these videos shows the temporal smoothness of pose estimates as well as their continuous alignment with the dancer's poses achieved by our method, which cannot be seen from the static depictions in the paper. Further, the pose and alignment quality can now be seen for the entire range of complexities associated with the poses assumed by the dancer throughout the video instead of with the selected few frames in the paper.

The videos we use to show our results are selected from the test set in the \textit{UID dataset}. The selected videos can be found at \url{https://drive.google.com/drive/folders/1X5K2U1Eq1QlcU8GmM_gHFVv75VkBzcoV?usp=sharing}. The right side of each video shows videos of skeletons representing 2D projections (2D poses) $\hat{p}_t$ of the estimated 3D poses $\hat{P}_t$ by themselves. These 2D poses are estimated using the estimated 3D-to-2D projection parameter $\omega^{*2D}$. To bring out the poses, they are shown from a closer and different viewpoint than used to capture the original video.
On the left side, we show the same 2D poses, using the viewpoint used to capture the original video, and overlaid on the original video frames. 
We can see that the skeletons align well with the complex movements of the dancers, such as spinning, Pointe (fully extended feet), and Tour Jeté (a high turning leap). Also, we can see that transitions between poses in adjacent frames are smooth, e.g., without large, abnormal displacements between the locations of the same joint in successive frames.

Finally, names of the recognized 3D movements $\hat{y}^e_t$ of each body part $e \in E$ are shown at the bottom of the video. The recognized movements $\hat{y}^e_t$ can be seen to well match the dancers' movements in the original video.

\section{Algorithms in Detail}
Algorithms \hyperref[alg:trackingSupplementary]{1*},
\hyperref[alg:2DposeEstSupplementary]{2*}, \hyperref[alg:3Dto2DSupplementary]{3*} and
\hyperref[alg:3DSupplementary]{4*} here are the detailed versions of Algorithms \ref{alg:tracking}, \ref{alg:2Dpose_est}, \ref{alg:3D_to_2D} and \ref{alg:3D}.
The movement and dance genre recognition is described in detail in Algorithm \ref{alg:MovementSupplementary} and \ref{alg:DanceClassifySupplementary}.

\begin{algorithm}[!ht]
\SetAlgoRefName{1*}
\SetAlgoLined
\textbf{Input}: a sequence of video frames $\{I_t\}^{T-1}_{t=0}$ \\
\textbf{Output}: a sequence of bounding boxes $\{(x_t^i, y_t^i, w_t^i, l_t^i)\}^{T-1}_{t=0}$ of the $i^{th}$ dancer \\
Initialization: select the bounding box $(x_0^i, y_0^i, w_0^i, l_0^i)$ of $N$ dancers to track by mouth \\
\While{new frame $I_t$ available}{
\For {$i^{th}$ dancer}{ 
Obtain $(x_t^i, y_t^i, w_t^i, l_t^i)$ by LDES approach \\
\If{not overlap with others}{
  $\tilde{h}$ $\leftarrow$ $h_t^i$ // Store histogram of $i^{th}$ dancer \\
  $\tilde{v}$ $\leftarrow$ $v_t^i$  // Store velocity of $i^{th}$ dancer \\
  }
\If{overlap happens \& tracking fails}{
  Estimate when overlap ends
  }
\If{overlap ends}{
  // Relocate the bounding box \\
  $\hat{k} = \argmax_k  (correlation(\tilde{h}, h^k))$ where $h^k$ is the histogram of the $k^{th}$ patch along the moving direction in the cone searching region \\
  $(x_t^i, y_t^i, w_t^i, l_t^i)$ $\leftarrow$ location of $\hat{k}^{th}$ patch
  }
}
}
\caption{Object Tracking}
\label{alg:trackingSupplementary}
\end{algorithm}

\begin{algorithm}[!ht]
\SetAlgoLined
\SetAlgoRefName{2*}
\textbf{Input}: a sequence of video frames $\{I_t\}^{T-1}_{t=0}$ and a sequence of bounding boxes $\{B_t^i\}^{T-1}_{t=0} = \{(x_t^i, y_t^i, w_t^i, l_t^i)\}^{T-1}_{t=0}$ of the $i^{th}$ dancer\\
\textbf{Output}: a sequence of poses $\{\hat{p}_t^i\}^{T-1}_{t=0}$ of the $i^{th}$ dancer \\
\While{new frame $I_t$ available}{
Estimate poses // Perform OpenPose\\
\For {$i^{th}$ dancer}{ 
Select $C$ poses $\{p_t^{i,c}\}^{C-1}_{c=0}$ overlapped with the bounding box $B_t^i$ \\
$\hat{c} = \argmax_c  (correlation(h_t^{i,c}, h_{t-1}^i))$ where $h_t^{i,c}$ is the histogram of the pose $p_t^{i,c}$\\
$\hat{p}_t^i$ $\leftarrow$ $p_t^{i,\hat{c}}$
}
}
\caption{Tracking Based 2D Pose Estimation}
\label{alg:2DposeEstSupplementary}
\end{algorithm}

\begin{algorithm}[!ht]
\SetAlgoRefName{3*}
\SetAlgoLined
\textbf{Input}: a sequence of 2D poses $\{p_t\}^{N-1}_{t=0}$ of a dancer \\
\textbf{Output}: a sequence of 3D poses $\{\tilde{P}_t\}^{N-1}_{t=0}$ of the dancer \\
Set the temporal window size to be $2\Delta$ \\
Denote total number of segments as $s=\floor*{\frac{N}{2\Delta}}$ \\
\For{$t = \Delta$ to $N-\Delta$}{
\For{k = 0 to $K-1$}{
Try new seed for DH parameters $\Lambda^k = \{\bm{\Theta}^k,\bm{d}^k,\bm{a}^k,\bm{\alpha}^k\}$ and perspective projection parameters $\omega^k=\{f^k,c^k\}$ \\
\For{$i = t-\Delta$ to $t+\Delta$}{
Generate 3D pose $\hat{P}^k_i=G(\Lambda^k)$ \\
Estimate 2D pose $\hat{p}_{i}^k=\Psi(\hat{P}_{i}^k; \omega^k)$ \\
Compute error $e^k_i = ||\hat{p}_{i}^k - p_{i}||^2_2$ \\
Optimize $\Lambda^{*k}, \omega^{*k}=\argmin_{\Lambda^k, \omega^k} e^k_i$ \\
Assign $\hat{P}_i^{*k} = G(\Lambda^{*k})$ \\
Update $\Lambda^{k} \leftarrow \Lambda^{*k}$ \\
Update $\omega^k \leftarrow  \frac{1}{i-t+\Delta} \sum_{l=t-\Delta}^{t} \omega_l^{*k}$ \\}
}
}
Select the seed $k^{*} = \argmin_{\tilde{k}}\sum_{i=t-\Delta}^{t+\Delta} e^{\tilde{k}}_i$ \\
Assign $\tilde{P}_t, \omega_t^{2D} \leftarrow \hat{P}_t^{*k^{*}}, \omega^{k^{*}}$
\caption{3D Pose Initialization}
\label{alg:3Dto2DSupplementary}
\end{algorithm}

\begin{algorithm}[!ht]
\SetAlgoRefName{4*}
\SetAlgoLined
\textbf{Input}: a sequence of video frames $\{I_t\}^{T-1}_{t=0}$, 2D poses $\{p_t\}^{T-1}_{t=0}$ and initial 3D poses $\{\tilde{P}_t\}^{T-1}_{t=0}$ of a dancer \\
\textbf{Output}: a sequence of estimated 3D poses $\{\hat{P}_t\}^{T-1}_{t=0}$ of the dancer \\
\While{new frame $I_{t}$ available}{
Estimate 3D pose $ \hat{P}_{t} = \Phi(p_{t}; \omega^{3D}) $ \\
Project to 2D pose $\hat{p}_{t} = \Psi(\hat{P}_{t}; \omega^{2D})$ \\
Compute loss $L = \alpha(|| \hat{p}_{t} - \hat{p}_{t-1} ||^2_2 + \beta || \hat{P}_{t} - \hat{P}_{t-1} ||^2_2) + || \hat{p}_{t} - p_{t} ||^2_2 + || \hat{P}_{t} - \tilde{P}_{t} ||^2_2$ \\
Update $\omega^{2D}$ $\leftarrow$ $\omega^{2D} - \eta \frac{\partial L}{\partial \omega^{2D}} $ \\
Update $\omega^{3D}$ $\leftarrow$ $\omega^{3D} - \eta \frac{\partial L}{\partial \omega^{3D}} $
}
\caption{3D Pose Estimation}
\label{alg:3DSupplementary}
\end{algorithm}

\begin{algorithm}[!ht]
\SetAlgoLined
\textbf{Input}: a sequence of poses $\{\bar{p}_t^e=\{(\bar{x}^j_t,\bar{y}^j_t)\}^{|J_e|}_{j=0}\}^{T-1}_{t=0}$ where $T$ denotes the total number of frames and $J_e$ denotes the set of body joints connected to the body part $e$; a sequence of corresponding movement labels $\{\tilde{y}^e_t\}^{T-1}_{t=0}$ of the body part $e$ \\
\textbf{Output}: predicted movement labels $\{\hat{y}^e_t\}^{T-1}_{t=0}$ \\
\For{$\text{epoch}=0$ \textbf{to} $N-1$}{
$\{\hat{y}_t^e\}^{T-1}_{t=0} = \text{LSTM}(\{\bar{p}_t^e\}^{T-1}_{t=0})$ \\
$L = \text{BCELoss}(\{\hat{y}_t^e\}^{T-1}_{t=0}, \{\tilde{y}^e_t\}^{T-1}_{t=0})$ \\
Update \text{LSTM} until converge
}
\caption{Movement Identification}
\label{alg:MovementSupplementary}
\end{algorithm}

\begin{algorithm}[!ht]
\SetAlgoLined
\textbf{Input}: a sequence of movement labels $\{\{\hat{y}^e_t\}^{|E|-1}_{e=0}\}^{T-1}_{t=0}$ of all the body parts $e \in E$; and the ground-truth dance genre label $g$ of the sequence \\
\textbf{Output}: predicted dance genre label $\hat{g}$ \\
\For{$\text{epoch}=0$ \textbf{to} $N-1$}{
$\hat{g} = \text{LSTM}(\{\{\hat{y}^e_t\}^{|E|-1}_{e=0}\}^{T-1}_{t=0})$ \\
$\tilde{L} = \text{CrossEntropyLoss}(\hat{g}, g)$ \\
Update \text{LSTM} until converge
}
\caption{Dance Classification}
\label{alg:DanceClassifySupplementary}
\end{algorithm}

\section{Implementation Details}
Table \ref{tab:DH_parameters} and \ref{tab:bounds} show the the values of the DH parameters, and the bounds of the joint rotation offset angles and bone length of our 34-DOF digital dancer model.

\begin{table*}[!ht]
\small
\centering
\begin{tabular}{p{0.4cm}p{1.2cm}lll|p{0.4cm}p{1.3cm}lll|p{0.4cm}p{1.3cm}lll}
	\toprule
	\textbf{Joint} & \textbf{$\Theta$} & \textbf{$d$} & \textbf{$a$} & \textbf{$\alpha$} & \textbf{Joint} & \textbf{$\Theta$} & \textbf{$d$} & \textbf{$a$} & \textbf{$\alpha$} & \textbf{Joint} & \textbf{$\Theta$} & \textbf{$d$} & \textbf{$a$} & \textbf{$\alpha$}  \\
	\midrule
    0 & $180+\theta_0$ & 0 & 0 & 90 & 12 & $90+\theta_{11}$ & $b_4 h$ & 0 & 90 & 24 & $0+\theta_{23}$ & 0 & 0 & 90  \\
    1 & $-90+\theta_0$ & 0 & 0 & 90 & 13 & $90+\theta_{12}$ & 0 & $0.6 b_4 h$ & 0 & 25 & $0+\theta_{24}$ & $b_8 h$ & 0 & -90 \\
	2 & $90+\theta_1$ & 0 & $b_0 h$ & -90 & 14 & $0+\theta_{13}$ & 0 & 0 & -90 & 26 & $-90+\theta_{25}$ & 0 & $0.1 b_8 h$ & 0 \\
	3 & $0+\theta_2$ & 0 & 0 & 90 & 15 & $-90+\theta_{14}$ & 0 & 0 & 90 &  27 & $0+\theta_{26}$ & 0 & 0 & -90 \\
	4 & $90+\theta_3$ & 0 & 0 & 90 & 16 & $90+\theta_{15}$ & $-b_3 h$ & 0 & 90 & 28 & $-90+\theta_{27}$ & 0 & 0 & 90 \\
	5 & $90+\theta_4$ & $b_1 h$ & 0 & 90 & 17 & $0+\theta_{16}$ & 0 & 0 & -90 & 29 & $0+\theta_{28}$ & $b_6 h$ & 0 & -90  \\
	6 & $90+\theta_5$ & 0 & 0 & 90 & 18 & $90+\theta_{17}$ & $-b_4 h$ & 0 & 90 & 30 & $90+\theta_{29}$ & 0 & $-b_7 h$ & 0  \\
	7 & $90+\theta_6$ & 0 & $b_1 h$ & 0 & 19 & $-90+\theta_{18}$ & 0 & $-0.6b_4 h$ & 0 & 31 & $0+\theta_{30}$ & 0 & 0 & 90 \\
	8 & $0+\theta_7$ & 0 & 0 & 90 & 20 & $0+\theta_{19}$ & 0 & 0 & -90 & 32 & $0+\theta_{31}$ & $-b_8 h$ & 0 & -90 \\
	9 & $90+\theta_8$ & 0 & 0 & 90 & 21 & $-90+\theta_{20}$ & 0 & 0 & 90 & 33 & $-90+\theta_{32}$ & 0 & $-0.1 b_8 h$ & 0 \\
	10 & $90+\theta_{9}$ & $b_3 h$ & 0 & 90 &  22 & $0+\theta_{21}$ & $-b_6 h$ & 0 & -90 &  &  &  &  & \\
	11 & $0+\theta_{10}$ & 0 & 0 & -90 & 23 & $90+\theta_{22}$ & 0 & $b_7 h$ & 0 &  &  &  &  & \\
	\bottomrule
\end{tabular}
\caption{The DH parameters $\Lambda = \{\bm{\Theta},\bm{d},\bm{a},\bm{\alpha}\}$ for the 34-DOF human model as shown in Figure \mbox{\ref{fig:human_model}}. The joint rotation angle $\bm{\Theta}$ along $z$ axis, the distance $\bm{d}$ along $z$ axis, and the offset distance $\bm{a}$ along $x$ axis are determined by the joint rotation offsets $\bm{\theta}=(\theta_0, ..., \theta_{32})$ and bone lengths $\bm{b}=(b_0, ..., b_6)$, where their bounds are defined in Table \mbox{\ref{tab:bounds}}.
}
\label{tab:DH_parameters}
\end{table*}

\begin{table*}[!ht]
\small
\centering
\setlength\tabcolsep{4pt}
\begin{tabular}{llllllllllllllllllllll}
	\toprule
    \textbf{Rotation} & $\theta_1$ & $\theta_2$ & $\theta_3$ & $\theta_7$ & $\theta_8$ & $\theta_9$ & $\theta_{10}$ & $\theta_{13}$ & $\theta_{14}$ & $\theta_{15}$ & $\theta_{16}$ & $\theta_{19}$ & $\theta_{20}$ & $\theta_{21}$ & $\theta_{22}$ & $\theta_{23}$ & $\theta_{26}$ & $\theta_{27}$ & $\theta_{28}$ & $\theta_{29}$ & $\theta_{30}$  \\
    \midrule
    $\textit{min}$ & $-\frac{\pi}{8}$ & $-\frac{\pi}{4}$ & $-\frac{\pi}{4}$ & $-\frac{\pi}{1.6}$ & $-\frac{\pi}{4}$ & $-\pi$ & $-\frac{\pi}{2}$ & $-\frac{\pi}{1.6}$ & $-\frac{\pi}{4}$ & $-\pi$ & $-\frac{\pi}{2}$ & $-\frac{\pi}{2}$ & $-\pi$ & $-\frac{\pi}{2}$ & 0 & $-\frac{\pi}{4}$ & $-\frac{\pi}{2}$ & $-\pi$ & $-\frac{\pi}{2}$ & 0 & $-\frac{\pi}{4}$\\
    $\textit{max}$ & $\frac{\pi}{8}$ & $\frac{\pi}{4}$ & $\frac{\pi}{4}$ & $\frac{\pi}{1.6}$ & $\frac{\pi}{1.6}$ & 0 & 0 & $\frac{\pi}{1.6}$ & $\frac{\pi}{1.6}$ & 0 & 0 & $\frac{\pi}{4}$ & $\frac{\pi}{1.3}$ & $\frac{\pi}{2}$ & $\frac{\pi}{1.3}$ & $\frac{\pi}{2}$ & $\frac{\pi}{4}$ & $\frac{\pi}{1.3}$ & $\frac{\pi}{2}$ & $\frac{\pi}{1.3}$ & $\frac{\pi}{2}$ \\
    \bottomrule
\end{tabular}
\begin{tabular}{llllllllll}
	\toprule
    \textbf{Bone} & neck $b_0$ &  head $b_1$ & shoulder $b_2$ & uparm $b_3$ & lowarm $b_4$ & hip $b_5$ & upleg $b_6$ & lowleg $b_7$ & toe $b_8$ \\
    \midrule
    $\textit{Average}$ & 0.25 & 0.08 & 0.06 & 0.17 & 0.17 & 0.04 & 0.21 & 0.21 & 0.04  \\
    $\textit{Std}$ & 0.05 & 0.05 & 0.05 & 0.05 & 0.05 & 0.05 & 0.05 & 0.05 & 0.05 \\
	\bottomrule
\end{tabular}
\caption{The bounds of the joint rotation  offset angles $\bm{\theta}=(\theta_0, ..., \theta_{32})$ and bone length ratios $\bm{b}=(b_0, ..., b_6)$ defined for our digital dancer model.}
\label{tab:bounds}
\end{table*}
\end{appendices}

\end{document}